\documentclass{article} 
\usepackage[final]{colm2025_conference}

\usepackage{microtype}

\usepackage{amsmath}
\usepackage{amssymb}
\usepackage{mathtools}
\usepackage{amsthm}

\usepackage{hyperref}
\usepackage{url}
\usepackage{wrapfig}
\usepackage{graphicx}
\usepackage{caption}
\usepackage{subcaption}
\usepackage{booktabs}
\usepackage{array}
\usepackage{multirow, multicol}
\usepackage{float}

\usepackage{lineno}

\definecolor{darkblue}{rgb}{0, 0, 0.5}
\hypersetup{colorlinks=true, citecolor=darkblue, linkcolor=darkblue, urlcolor=darkblue}

\title{BlockFFN: Towards End-Side Acceleration-Friendly Mixture-of-Experts with Chunk-Level Activation Sparsity}

\author{Chenyang Song\footnotemark[1], \ Weilin Zhao\footnotemark[1], \ Xu Han\footnotemark[2], \ Chaojun Xiao, Yingfa Chen,\\
\textbf{Yuxuan Li, Zhiyuan Liu\footnotemark[2], \ Maosong Sun}\\
Dept. of Comp. Sci. \& Tech., Institute for AI, Tsinghua University, Beijing, China\\
\texttt{\{scy22,zwl23\}@mails.tsinghua.edu.cn, \{han-xu,liuzy\}@tsinghua.edu.cn}
}

\begin{document}

\ifcolmsubmission
\linenumbers
\fi

\maketitle

\renewcommand{\thefootnote}{\fnsymbol{footnote}}
\footnotetext[1]{Equal Contributions.}
\footnotetext[2]{Corresponding Authors.}
\renewcommand{\thefootnote}{\arabic{footnote}}
\setcounter{footnote}{0}

\begin{abstract}
To alleviate the computational burden of large language models (LLMs), architectures with activation sparsity, represented by mixture-of-experts (MoE), have attracted increasing attention.
However, the non-differentiable and inflexible routing of vanilla MoE hurts model performance.
Moreover, while each token activates only a few parameters, these sparsely-activated architectures exhibit low chunk-level sparsity, indicating that the union of multiple consecutive tokens activates a large ratio of parameters.
Such a sparsity pattern is unfriendly for acceleration under low-resource conditions (e.g., end-side devices) and incompatible with mainstream acceleration techniques (e.g., speculative decoding).
To address these challenges, we introduce a novel MoE architecture, BlockFFN, as well as its efficient training and deployment techniques. 
Specifically, we use a router integrating ReLU activation and RMSNorm for differentiable and flexible routing.
Next, to promote both token-level sparsity (TLS) and chunk-level sparsity (CLS), CLS-aware training objectives are designed, making BlockFFN more acceleration-friendly. 
Finally, we implement efficient acceleration kernels, combining activation sparsity and speculative decoding for the first time.
The experimental results demonstrate the superior performance of BlockFFN over other MoE baselines, achieving over 80\% TLS and 70\% 8-token CLS. Our kernels achieve up to 3.67$\times$ speedup on real end-side devices than dense models.
All codes and checkpoints are available publicly\footnote{\url{https://github.com/thunlp/BlockFFN}}.


\end{abstract}

\section{Introduction}

To reduce the high costs of training and deploying large language models (LLMs), various efficient LLM architectures are proposed~\citep{wan2023efficient}.
A popular paradigm is designing architectures with \textbf{activation sparsity}, indicating that a considerable part of LLM parameters contribute weakly to LLM outputs given specific inputs, and thus can be skipped (i.e., not activated) in the forward and backward computation. Mixture-of-experts (MoE) is an outstanding representative and has been adopted by many recent models such as Mixtral-8$\times$22B~\citep{jiang2024mixtral} and DeepSeek-V3~\citep{liu2024deepseek}. Based on MoE, techniques including load balancing~\citep{wang2024auxiliary} and expert parallelism~\citep{he2021fastmoe} are adopted to achieve remarkable efficiency on cloud-side servers.




However, few efforts explore sparsely-activated architectures under low-resource conditions (e.g., end-side devices), where it is challenging to deploy huge MoE models and highly-distributed frameworks with expert parallelism. For end-side MoE models, which generally serve only a few users, some typical issues are not required to be considered (e.g., load balancing, see Appendix~\ref{sec:load-balancing}), while raising the following two challenges.



\textbf{Performance compromise caused by imperfect routing.}\quad
Existing MoE models generally compromise performance due to two significant routing drawbacks: non-differentiability and inflexibility. Specifically, most mainstream MoE models adopt a TopK router~\citep{fedus2022switch,jiang2024mixtral} with discrete and non-differentiable computation. Consequently, only activated parameters have complete gradients and are well updated at each step, which harms the convergence efficiency of MoE models~\citep{liu2024grin}. Moreover, TopK makes each token activate the same number of experts, enforcing an inflexible activation pattern, which may weaken model performance~\citep{huang2024harder}. Few works can well alleviate both drawbacks, see Section~\ref{sec:works-arch-sparsity}.

\textbf{Acceleration unfriendliness caused by low chunk-level sparsity (CLS).}\quad
To make a sparsely-activated architecture more friendly for acceleration, just increasing the ratio of weakly-contributed experts for each token, namely the token-level sparsity (TLS), is not enough. Instead, the ratio of weakly-contributed experts for multiple consecutive tokens, namely the chunk-level sparsity (CLS), is critical for practical acceleration.

\begin{figure}[t]
\begin{subfigure}[b]{0.325\textwidth}
    \centering
    \includegraphics[width=\linewidth]{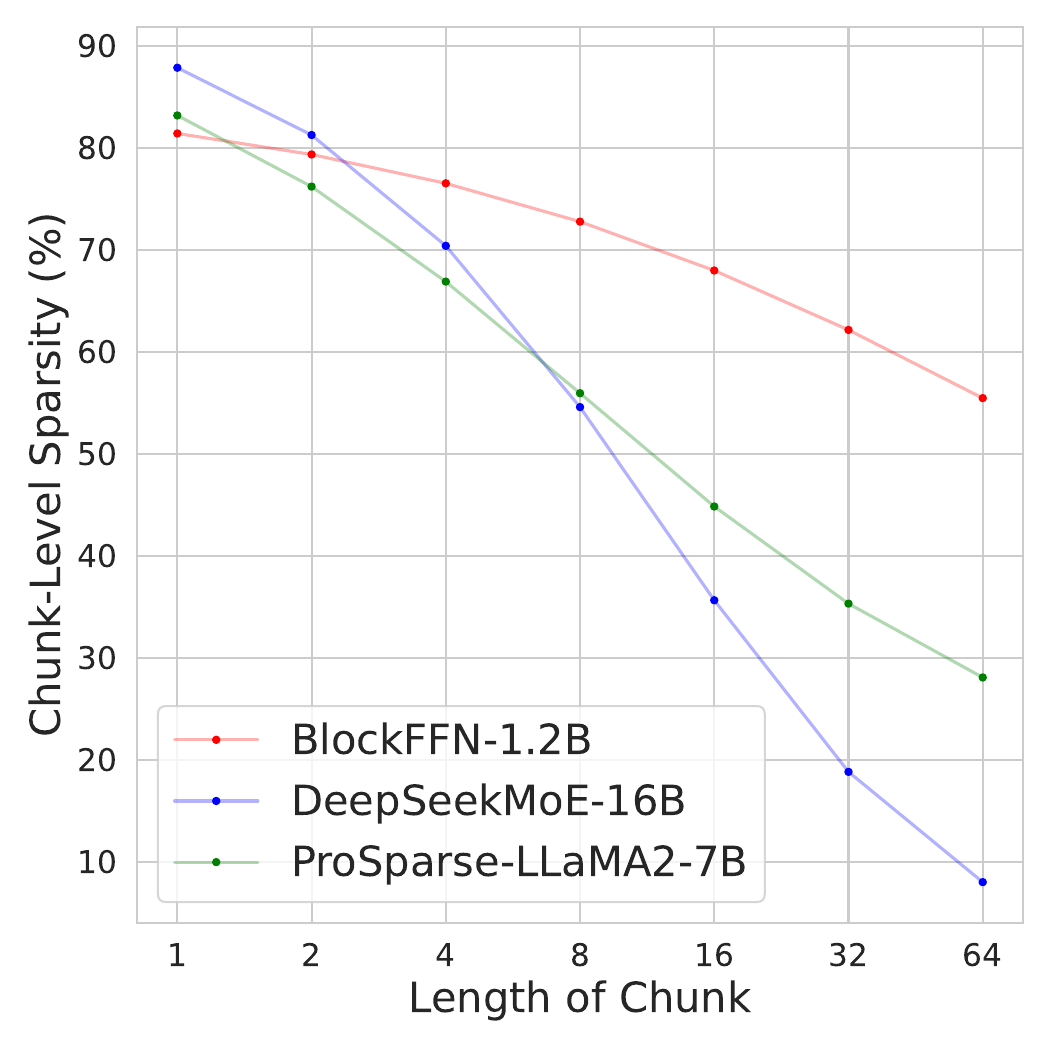}
    \captionsetup{skip=0.1em}
    \caption{}
    \label{fig:ssr-collapse}
\end{subfigure}
\hfill
\begin{subfigure}[b]{0.325\textwidth}
    \centering
    \includegraphics[width=\linewidth]{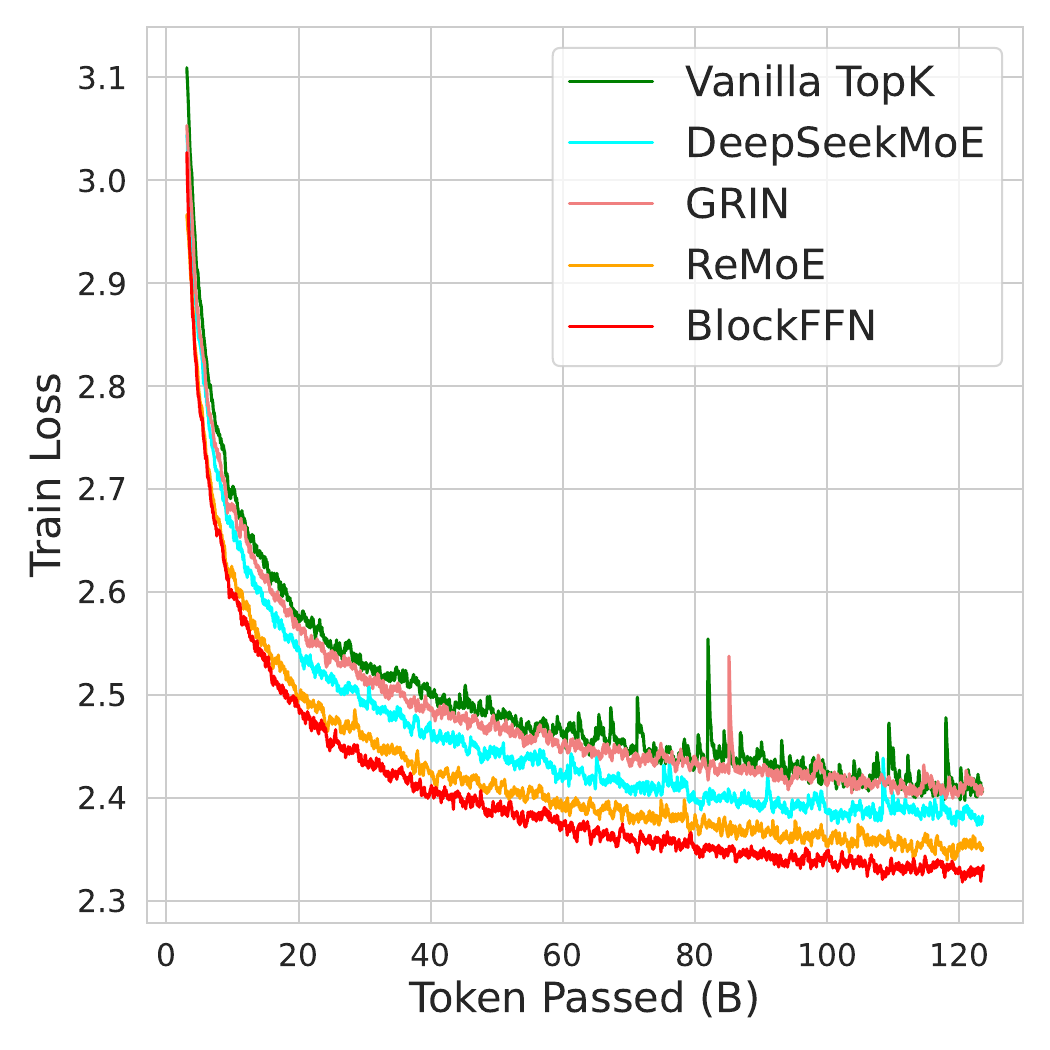}
    \captionsetup{skip=0.1em}
    \caption{}
    \label{fig:train-curve-12b}
\end{subfigure}
\hfill
\begin{subfigure}[b]{0.325\textwidth}
    \centering
    \includegraphics[width=\linewidth]{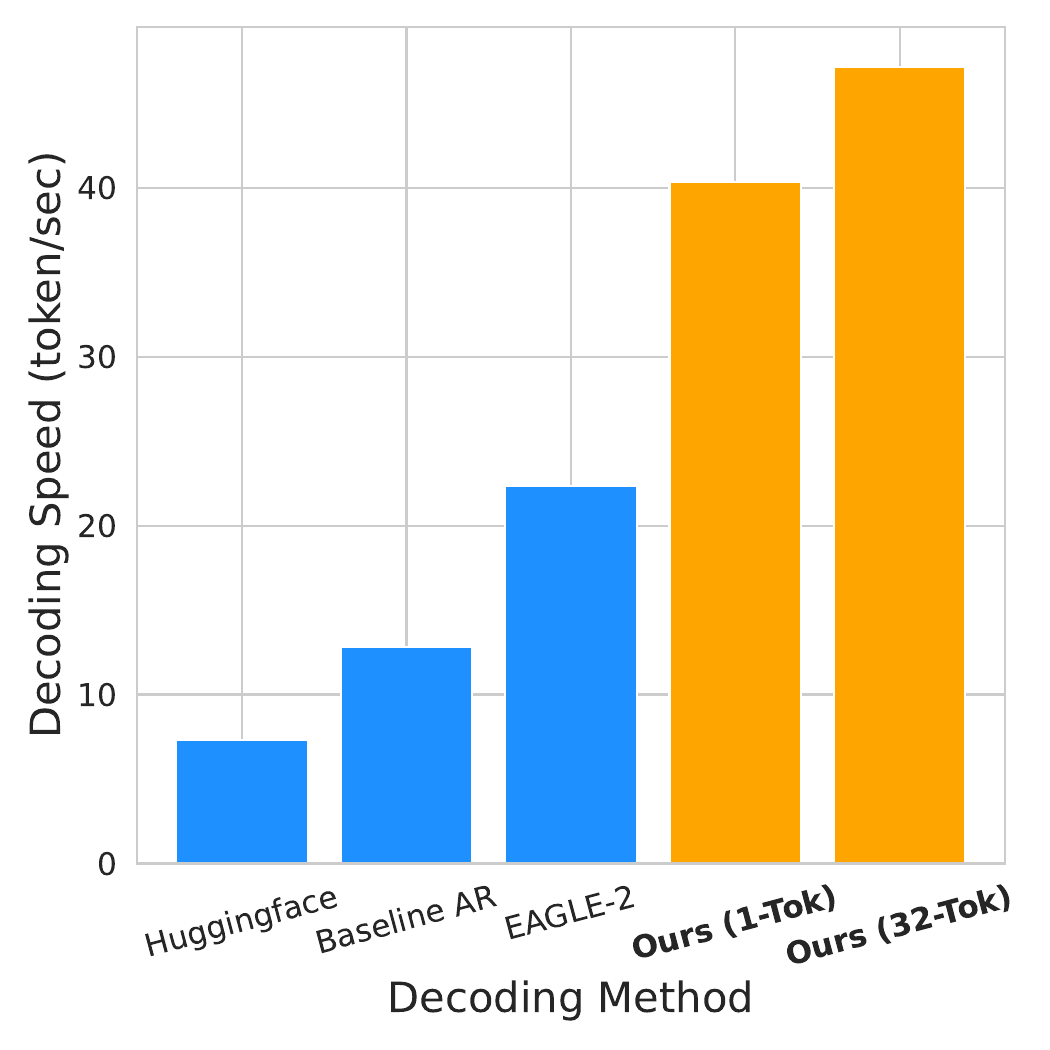}
    \captionsetup{skip=0.1em}
    \caption{}
    \label{fig:speedup-bar}
\end{subfigure}
\vspace{-0.8em}
\caption{(a) For models with high TLS (except BlockFFN-1.2B), CLS quickly collapses to a lower level as a chunk contains more consecutive tokens. (b) Smoothed training curves of the BlockFFN-1.2B and other MoE baselines with the same total and activated parameters. (c) The speed of different decoding methods, where ``Ours (1-Tok)'' and ``Ours (32-Tok)'' are our token-level and chunk-level sparsity-based acceleration kernels, respectively.}
\vspace{-1.5em}
\end{figure}

Specifically, a low CLS can eliminate the value of activation sparsity when combined with speculative decoding, a mainstream acceleration method that requires LLMs to process multiple consecutive tokens at the same time~\citep{leviathan2023fast}. Besides, other important resource-saving techniques, such as offloading, also become more challenging to implement due to the large differences in activation patterns within a specific chunk, leading to frequent GPU-CPU communication overheads. Unfortunately, existing works mainly focus on improving TLS~\citep{mirzadeh2023relu,song2024turbo,song2024prosparse}, but low CLS values still exist in most sparse architectures (see Figure~\ref{fig:ssr-collapse}).


To address the above challenges, we introduce the following novel MoE architecture, as well as its training techniques and efficient end-side deployment.

For model architectures, we propose \textbf{BlockFFN}, a novel MoE paradigm that minimizes performance compromise. Specifically, its router module integrates ReLU and RMSNorm. ReLU computes differentiable and flexible activation patterns, enabling each token to determine the number of activated experts adaptively. RMSNorm generates learnable magnitudes of activation values. This separation of activation patterns and magnitudes alleviates the disturbance on activation magnitudes induced by regularization (e.g., the shrinkage of activation magnitudes caused by L1~\citep{rajamanoharan2024improving}), as regularization applies solely to the ReLU activation pattern. Through experiments, we demonstrate the better performance of BlockFFN compared to other MoE variants (Figure~\ref{fig:train-curve-12b}, Table~\ref{tab:moe-ppl} and~\ref{tab:moe-eval}).

For training techniques, we introduce \textbf{CLS-aware training objectives} to improve the CLS of BlockFFN, which can make BlockFFN more friendly for acceleration, including the activation locality loss and the chunk sparsification loss. The former is to increase the similarity of activation patterns between neighbor tokens, helping reduce the gap between TLS and CLS. The latter is to increase the overall sparsity level. While existing sparsification objectives such as L1~\citep{song2024prosparse} are applied to each token independently, our chunk sparsification loss directly minimizes the probability that a specific expert is activated by at least one token within the chunk. In experiments, we obtain average TLS values higher than 80\% and 8-token CLS values higher than 70\% (Table~\ref{tab:moe-ppl}).

For end-side deployment, we implement \textbf{efficient acceleration kernels} for BlockFFN, combining activation sparsity and speculative decoding for the first time, and demonstrate its practical effectiveness on real end-side devices such as NVIDIA Jetson Orin NX. 
To enhance the efficiency of verifying multiple tokens in speculative sampling, we leverage the high activation similarity across multiple tokens induced by the high CLS level. This enables merging memory accesses to the same expert across different tokens, reducing the memory access volume to the union of experts activated by these tokens (Figure~\ref{fig:blockffn-kernel}).
Additionally, we implement the kernels based on CUTLASS~\citep{Thakkar_CUTLASS_2023} and utilize tensor cores to boost computational efficiency.
Overall, the kernel achieves an acceleration ratio of 3.67$\times$, compared to the baseline auto-regressive (AR) decoding (Figure~\ref{fig:speedup-bar} and Table~\ref{tab:speedup-overall}).

\section{Preliminaries and Related Works}

\begin{table*}[t]
    \small
    \centering
    \setlength{\tabcolsep}{0.5em}
    \scalebox{0.87}{
    \begin{tabular}{lll}
    \toprule
        Model & Activation values $\mathbf{A}(\mathbf{x})$ & Single expert outputs $E_i(\mathbf{x})$ \\
        \midrule
        \multicolumn{3}{c}{Neuron-level activation sparsity} \\
        \midrule
        \multirow{2}{*}{Non-gated FFN} & $\sigma(\mathbf{W}_{up}^T\mathbf{x})$ & $\mathbf{w}_{down}$ \\
        & $\mathbf{W}_{up}\in\mathbb{R}^{d_h\times N_e}$ & $\mathbf{w}_{down}\in\mathbb{R}^{d_h}$ \\[1ex]

        Gated FFN & $\sigma(\mathbf{W}_{gate}^T\mathbf{x})$ & $(\mathbf{w}_{up}^T\mathbf{x})\cdot\mathbf{w}_{down}$ \\
        \citep{shazeer2020glu} & $\mathbf{W}_{gate}\in\mathbb{R}^{d_h\times N_e}$ & $\mathbf{w}_{up},\mathbf{w}_{down}\in\mathbb{R}^{d_h}$ \\[1ex]

        dReLU & $\mathrm{ReLU}(\mathbf{W}_{gate}^T\mathbf{x})$ & $\mathrm{ReLU}(\mathbf{w}_{up}^T\mathbf{x})\cdot\mathbf{w}_{down}$ \\
        \citep{song2024turbo} & $\mathbf{W}_{gate}\in\mathbb{R}^{d_h\times N_e}$ & $\mathbf{w}_{up},\mathbf{w}_{down}\in\mathbb{R}^{d_h}$ \\
        \midrule
        \multicolumn{3}{c}{Block-level activation sparsity} \\
        \midrule
        Vanilla TopK MoE & $\mathrm{TopK}(\mathrm{Softmax}(\mathbf{W}_{router}^T\mathbf{x}))$ & \\
        \citep{jiang2024mixtral} & $\mathbf{W}_{router}\in\mathbb{R}^{d_h\times N_e}$ & \\[1ex]

        DeepSeek-V1/V2 & $[\mathbf{1}_{share},\mathrm{TopK}(\mathrm{Softmax}(\mathbf{W}_{router}^T\mathbf{x}))]$ & \\
        \citep{dai2024deepseekmoe} & $\mathbf{W}_{router}\in\mathbb{R}^{d_h\times (N_e-N_{share})}$ & \\[1ex]

        DeepSeek-V3 & $[\mathbf{1}_{share},\mathrm{Norm}(\mathrm{TopK}(\mathrm{Sigmoid}(\mathbf{W}_{router}^T\mathbf{x})))]$ & \\
        \citep{liu2024deepseek} & $\mathbf{W}_{router}\in\mathbb{R}^{d_h\times (N_e-N_{share})}$ & $\mathbf{W}_{down}^T[\sigma(\mathbf{W}_{gate}^T\mathbf{x})\odot\mathbf{W}_{up}^T\mathbf{x}]$ \\[1ex]

        DynamicMoE & $\mathrm{TopP}(\mathrm{Softmax}(\mathbf{W}_{router}^T\mathbf{x}))$ & $\mathbf{W}_{gate},\mathbf{W}_{up}\in\mathbb{R}^{d_h\times d_e}$ \\
        \citep{huang2024harder} & $\mathbf{W}_{router}\in\mathbb{R}^{d_h\times N_e}$ & $\mathbf{W}_{down}\in\mathbb{R}^{d_e\times d_h}$ \\[1ex]

        GRIN & $\mathrm{SparseMixer}(\mathrm{Softmax}(\mathbf{W}_{router}^T\mathbf{x}))$ & \\
        \citep{liu2024grin} & $\mathbf{W}_{router}\in\mathbb{R}^{d_h\times N_e}$ & \\[1ex]

        ReMoE & $\mathrm{ReLU}(\mathbf{W}_{router}^T\mathbf{x})$ & \\
        \citep{wang2024remoe} & $\mathbf{W}_{router}\in\mathbb{R}^{d_h\times N_e}$ & \\[1ex]

        \midrule

        \textbf{BlockFFN} (Ours) & $\mathrm{RMSNorm}(\mathrm{ReLU}(\mathbf{W}_{router}^T\mathbf{x}))$ & $\mathbf{W}_{down}^T\sigma(\mathbf{W}_{up}^T\mathbf{x})$ \\[1ex]
        & $\mathbf{W}_{router}\in\mathbb{R}^{d_h\times N_e}$ & $\mathbf{W}_{up}\in\mathbb{R}^{d_h\times d_e},\mathbf{W}_{down}\in\mathbb{R}^{d_e\times d_h}$ \\
    \bottomrule
    \end{tabular}}
    \caption{Comparison between different architectures with activation sparsity. $\sigma$ denotes an activation function (e.g., ReLU, Swish, GELU). $d_e$ is the intermediate dimension of each block-level expert. For simplicity, the expert index $i$ is omitted in notations of $E_i(\mathbf{x})$.}
    \label{tab:arch-sparsity}
    \vspace{-1.5em}
\end{table*}


\subsection{Architectures with Activation Sparsity} \label{sec:works-arch-sparsity}

To reduce the computation expenses of LLMs, various inference acceleration methods are proposed. Quantization~\citep{xiao2023smoothquant,yao2023comprehensive,shao2023omniquant} and distillation~\citep{gu2023knowledge,hsieh2023distilling} compress LLMs by using low bit-widths and transferring knowledge into smaller models, respectively. Weight pruning~\citep{ma2023llm,sun2023simple,frantar2023sparsegpt,xia2023sheared} reduces FLOPs by removing weakly-contributed parameters (regardless of inputs). Speculative decoding~\citep{li2024eagle,cai2024medusa,zhao2024ouroboros} uses a smaller model to generate multiple candidate tokens and lets the LLM itself verify these tokens in parallel.

Besides the above post-training methods, efficient architectures with activation sparsity can also effectively reduce the computation overhead of LLMs~\citep{xue2024powerinfer,zhang2024exploring,liu2024deepseek}. Specifically, in sparsely-activated architectures, a considerable part of the parameters contribute weakly to the model outputs given specific inputs.

In this work, we mainly focus on the activation sparsity within FFN layers.
Typically, a sparsely-activated FFN with hidden dimension $d_h$ can be written in a unified MoE format:
\begin{small}
\begin{equation}
    \begin{aligned}
    \label{eq:moe-format}
    \text{FFN}(\mathbf{x}) = \sum_{i=1}^{N_e} A_i(\mathbf{x}) \cdot E_i(\mathbf{x}),\quad
    \mathbf{A}(\mathbf{x})=[A_1(\mathbf{x}),A_2(\mathbf{x}),...,A_{N_e}(\mathbf{x})],
    \end{aligned}
\end{equation}
\end{small}
\hspace{-\fontdimen2\font}where $\mathbf{x}\in\mathbb{R}^{d_h}$ is the input hidden state, and $N_e$ is the number of experts. $A_i(\mathbf{x})\in\mathbb{R}$ and $E_i(\mathbf{x})\in\mathbb{R}^{d_h}$ denote the $i$-th activation value and expert outputs, respectively.
If some $A_i(\mathbf{x})$ is zero or a low value, the corresponding expert $E_i$ is weakly-contributed. \textbf{Token-level sparsity $TLS$} denotes the average ratio of weakly-contributed experts for a single token, while \textbf{chunk-level sparsity $CLS_L$} is the ratio of experts contributing weakly to all tokens within a consecutive chunk of length $L$.
Based on the granularity of experts, LLM architectures with activation sparsity can be divided into the following two categories.

\textbf{Neuron-level activation sparsity} commonly exists in mainstream LLMs~\citep{li2022lazy}, where each expert is composed of a single neuron, i.e., single columns or rows within the FFN parameter matrices. For example, LLaMA2-7B is estimated to have about 70\% TLS~\citep{zhang2024relu}. However, as every single neuron is generally too small to be a memory access unit (i.e., $<1\mathrm{MB}$ in BF16), most neuron-level sparse LLMs have bad memory locality. This makes it difficult to realize practical acceleration due to large IO overheads.


\textbf{Block-level activation sparsity} indicates that each expert is composed of multiple neurons or MLP modules. Represented by MoE~\citep{fedus2022switch}, such architectures are currently the mainstream solution for sparsity-based acceleration due to their better memory locality.

Nevertheless, the routing strategies of many MoE models, especially TopK routers, have non-differentiable and inflexible activation patterns, which limit model performance~\citep{luo2024sparsing}. Works such as GRIN~\citep{liu2024grin}, ReMoE~\citep{wang2024remoe}, and DynamicMoE~\citep{huang2024harder} try to address these issues. We list and compare several representative architectures of activation sparsity in Table~\ref{tab:arch-sparsity}. There are also special methods based on direct expert merging (e.g., SMEAR~\citep{muqeeth2023soft} and Lory~\citep{zhong2024lory}) that cannot be naively expressed as Equation~\ref{eq:moe-format}.

In this work, BlockFFN absorbs the merits of both categories. With good memory locality of block-level experts, it adopts a ReLU-activated router (common in neuron-level settings), with activation values scaled by RMSNorm (i.e., the architectural difference from ReMoE).

\subsection{Acceleration with Activation Sparsity}

For neuron-level architectures, due to the relatively bad memory locality, designing tailored acceleration frameworks is complicated. Deja Vu~\citep{liu2023deja} and PowerInfer~\citep{song2023powerinfer} utilize activation predictors to forecast the activation values, thus reducing IO overheads. PowerInfer-2~\citep{xue2024powerinfer} introduces complex IO pipelines and neuron caches to promote higher speedup on specific smartphones. However, these all risk potentially inaccurate inference due to the imperfect performance of activation predictors.

Block-level architectures have relatively more available frameworks. FastMoE~\citep{he2021fastmoe} and Tutel~\citep{hwang2023tutel} mainly focus on distributed training or inference with multiple GPUs working concurrently, while MegaBlocks~\citep{gale2023megablocks} emphasizes the large-batch training of MoE. However, few of them are tailored for deploying MoE on end-side devices, where it is generally impractical to adopt a distributed implementation, and the service requirements shrink to small-batch inference for individual users. Under end-side conditions, sparsity-based acceleration will face different challenges.


As far as we know, \textbf{we present the first work to address the acceleration combining activation sparsity and speculative decoding}. Specifically, we improve the chunk-level sparsity of models through CLS-aware training objectives, making BlockFFN more friendly for sparsity-based acceleration and speculative decoding. Moreover, our acceleration kernels are well applicable to end devices and have remarkable effectiveness.

\section{Methodology}

In this section, we first introduce the overall architecture of BlockFFN (Section~\ref{sec:blockffn-arch}) and CLS-aware training objectives (Section~\ref{sec:training-objective}). Then, the acceleration kernels are introduced in Section~\ref{sec:acceleration-kernel}, combining activation sparsity and speculative decoding for the first time.

\subsection{BlockFFN Architecture} \label{sec:blockffn-arch}

\paragraph{Expert modules} Considering the better memory locality of block-level activation sparsity, we make each BlockFFN expert an MLP with an activation function:
\begin{small}
\begin{equation}
    \begin{aligned}
    \label{eq:blockffn-expert}
    E_i(\mathbf{x}) = \mathbf{W}_{down}^{(i)T}\ \mathrm{Swish}(\mathbf{W}_{up}^{(i)T}\mathbf{x}),
    \end{aligned}
\end{equation}
\end{small}
\hspace{-\fontdimen2\font}where $i$ is the expert index, and $\mathbf{W}_{up}^{(i)}\in\mathbb{R}^{d_h\times d_e}, \mathbf{W}_{down}^{(i)}\in\mathbb{R}^{d_e\times d_h}$ are learnable weights.

Following DeepSeekMoE~\citep{liu2024deepseek}, we use fine-grained expert segmentation to increase flexibility, namely $d_e<<d_h$. We specifically add a Swish activation~\citep{ramachandran2017searching} to increase the nonlinearity. Notably, we choose a vanilla non-gated MLP for experts instead of the more popular gated variant~\citep{dauphin2017language,shazeer2020glu}, as we find that a gated MLP can destroy the router sparsity (See Appendix~\ref{sec:gated-expert}).


\paragraph{Router module} BlockFFN adopts a linear router with ReLU activation instead of TopK. As a common activation function in neuron-level sparse LLMs, ReLU is fully differentiable and can generate sparser activation patterns than other common activations (e.g., Swish)~\citep{luo2024sparsing}. Moreover, ReLU allows each token to adaptively activate different numbers of experts. This alleviates the inflexibility issue of conventional TopK routing.

On the other hand, as one major difference from ReMoE~\citep{wang2024remoe}, we add an RMSNorm layer~\citep{zhang2019root} after ReLU:
\begin{small}
\begin{equation}
    \begin{aligned}
    \label{eq:blockffn-router}
    \mathbf{A}^0(\mathbf{x})=\mathbf{W}_{router}^T\mathbf{x},\quad 
    \mathbf{A}^1(\mathbf{x})=\mathrm{ReLU}(\mathbf{A}^0(\mathbf{x})),\quad
    \mathbf{A}(\mathbf{x}) = \mathrm{RMSNorm}(\mathbf{A}^1(\mathbf{x})),
    \end{aligned}
\end{equation}
\end{small}
\hspace{-\fontdimen2\font}where $\mathbf{W}_{router}$ is learnable parameters. Such a design makes the magnitude of activation values adaptively learned through RMSNorm, indicating better flexibility than vanilla softmax. Besides, RMSNorm separates the ReLU activation pattern $\mathbf{A}^1(\mathbf{x})$ from the final activation value $\mathbf{A}(\mathbf{x})$. This alleviates the disturbance on activation magnitudes by a direct regularization, which may hurt performance~\citep{rajamanoharan2024improving} (Section~\ref{sec:rmsnorm-disturbance}).

\subsection{CLS-Aware Training Objectives} \label{sec:training-objective}

The low chunk-level sparsity (CLS) is one important obstacle to fully leveraging activation sparsity in practical acceleration, especially under conditions where multiple consecutive tokens are processed in parallel (e.g., speculative decoding).
The improvement of CLS involves two important aspects: (1) how to promote activation locality; (2) how to promote higher overall sparsity. We propose two respective training objectives.

\paragraph{Activation locality loss}

Activation locality refers to the similarity of activation patterns between neighbor tokens, which also indicates the gap between TLS and CLS. To promote this property, we introduce the activation locality loss as an additional training objective:
\begin{small}
\begin{equation}
    \begin{aligned}
    \label{eq:activation-locality}
    \mathbf{A}_s^0(\mathbf{x})=\mathrm{LeftShift}(\mathbf{A}^0(\mathbf{x})),\quad
    \mathcal{L}_{al}=\mathrm{BCE}[\sigma(\alpha\cdot\mathbf{A}^0(\mathbf{x})),\sigma(\alpha\cdot\mathbf{A}_s^0(\mathbf{x}))],
    \end{aligned}
\end{equation}
\end{small}
\hspace{-\fontdimen2\font}where $\sigma$ and $\alpha$ denote the sigmoid function and the sharpness hyper-parameter, respectively.
We approximate the activation pattern through a sharp sigmoid function applied on $\mathbf{A}^0(\mathbf{x})$. $\mathrm{LeftShift}$ operator left-shifts a tensor in the sequence dimension, and finally, the binary cross entropy $\mathrm{BCE}$ minimizes the gap between the soft activation patterns of neighbor tokens.

\paragraph{Chunk sparsification loss}

Despite the increase in activation locality, practical acceleration cannot be achieved without a considerable reduction in computation, which relies on a high sparsity level. Conventionally, L1~\citep{song2024prosparse} and router entropy~\citep{huang2024harder} are both effective methods to improve sparsity, but they are applied independently to each token and cannot directly optimize the chunk-level sparsity.

Therefore, we design the chunk sparsification loss, which directly minimizes the chunk-level sparsity of a chunk with $L$ consecutive tokens. Suppose $p_{ik}$ is the probability of the $i$-th expert activated by the $k$-th token, while $\sum_{i=1}^{N_e} p_{ik}=1$. The loss is the average probability that the $i$-th expert is activated by at least one token within this chunk (i.e., $\mathcal{P}^i_{act}$):
\begin{small}
\begin{equation}
    \begin{aligned}
    \label{eq:chunk-sparsification}
    [p_{ik}]_{i=1}^{N_e}=\mathrm{Norm}(\mathbf{A}^1(\mathbf{x})),\quad
    \mathcal{P}^i_{act}=1-\exp(\sum_{k=1}^L\ln(1-p_{ik})),\quad
    \mathcal{L}_{cs}=\frac{1}{N_e}\sum_{i=1}^{N_e}\mathcal{P}^i_{act},
    \end{aligned}
\end{equation}
\end{small}
\hspace{-\fontdimen2\font}where $\mathbf{A}^1(\mathbf{x})\in\mathbb{R}^{N_e}$ specifically denotes the ReLU activation pattern of the $k$-th token, and $\mathrm{Norm}$ operator normailizes it in the expert dimension.

The overall training objectives are computed by: $\mathcal{L}_{total}=\mathcal{L}_{lm}+\lambda_{al}\mathcal{L}_{al}+\lambda_{cs}\mathcal{L}_{cs}$, where $\lambda_{al}$ and $\lambda_{cs}$ are corresponding factors. We introduce an \textbf{adaptive factor scheduler} to adaptively determine $\lambda_{cs}$ according to the dynamics of $\mathcal{L}_{cs}$, see Appendix~\ref{sec:adaptive-scheduler}.

\subsection{Acceleration Kernels} \label{sec:acceleration-kernel}

\begin{figure*}
    \centering
    \includegraphics[width=\linewidth]{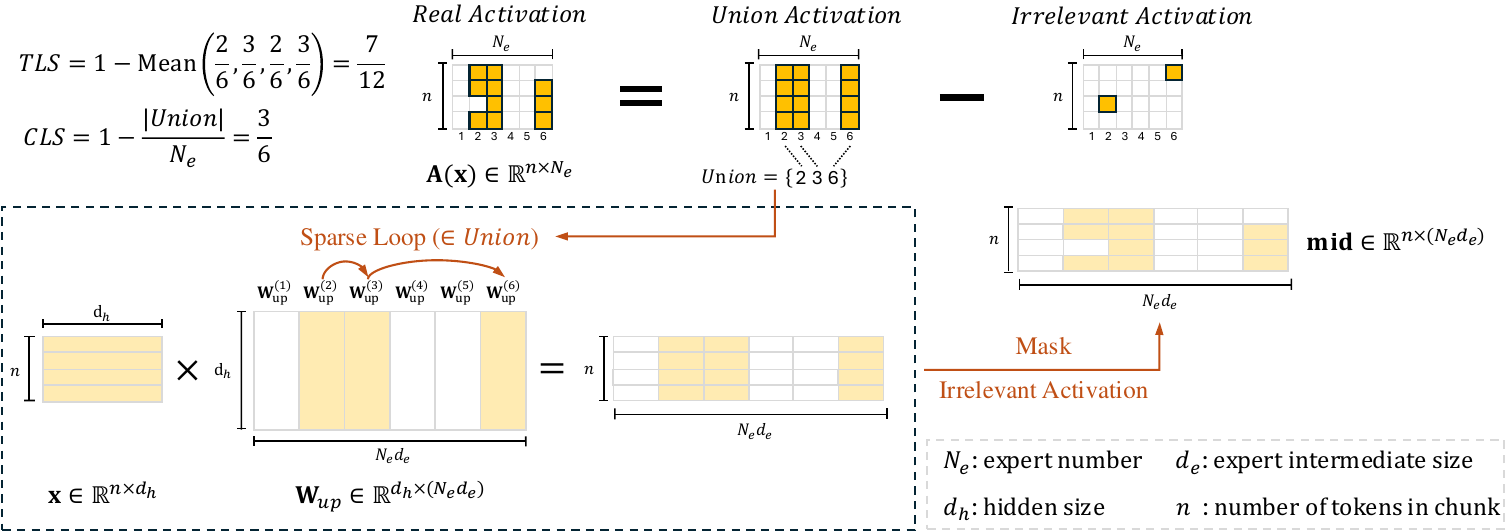}
    \caption{The overall framework of our acceleration kernels (the up projection part), which combines speculative decoding and chunk-level sparsity for higher efficiency. The down projection part has a similar implementation.}
    \label{fig:blockffn-kernel}
    \vspace{-1.5em}
\end{figure*}

We implement acceleration kernels for BlockFFN, which are applicable to end-side devices and effectively combine chunk-level sparsity and speculative decoding.

Specifically, during the speculative sampling process, the draft model proposes $n$ draft tokens. When BlockFFN verifies these tokens, the router activation values are denoted as $\textbf{A}(\mathbf{x}) \in \mathbb{R}^{n \times N_e} $, while the index union of activated experts for these $n$ tokens is $Union(\mathbf{x})$. Due to BlockFFN's high CLS level, the size of $Union(\mathbf{x}) $ only accounts for a small ratio to the total expert number. Therefore, by only involving the experts in $Union(\mathbf{x}) $ for computation, memory access is reduced and sparsity-based acceleration can be achieved in verification.

However, different experts may be activated by different subsets of tokens, which is not friendly for hardware parallelization. To address this issue, we leverage the characteristic that CLS and TLS values of BlockFFN are relatively close, indicating that each expert in $Union(\mathbf{x})$ is activated by the vast majority of tokens. Therefore, we only need to precompute all $n$ tokens for every activated expert for better GPU utilization, and subsequently discard computations induced by irrelevant activations.
Specifically, for up projection, the hidden states of all $n$ tokens participate in the matrix multiplication with the experts in $Union(\mathbf{x})$, yielding an intermediate result $\textbf{mid}$, as illustrated in Figure~\ref{fig:blockffn-kernel}. Finally, we apply a mask based on the sparse pattern of $\textbf{A}(\mathbf{x})$ to remove irrelevant activations. Similar sparse computation is also conducted for down projection.

The matrix multiplication kernel is modified based on CUTLASS GEMM~\citep{Thakkar_CUTLASS_2023}, where we modify the outer loop of the up projection and the inner loop of the down projection to only scan through those activated experts in $Union(\mathbf{x})$, see Appendix~\ref{sec:kernel-detail}. To match the requirements of CUDA Tensor Core, we set the number of draft tokens $n$ to 32.

\section{Experiments}

\subsection{Architecture Rationality}

\subsubsection{Overall Results} \label{sec:overall-result}

To demonstrate the rationality of our architecture, we conduct experiments by comparing BlockFFN with multiple sparsely-activated architectures: Vanilla TopK MoE, DeepSeekMoE (DSMoE)~\citep{dai2024deepseekmoe}, GRIN~\citep{liu2024grin}, and ReMoE~\citep{wang2024remoe} (see Appendix~\ref{sec:experiment-setting}).
To ensure fairness, we keep consistent settings for attention layers and MoE experts (i.e., the number and intermediate dimension of experts) throughout baselines and BlockFFN. Besides, all settings (within each scale) have close parameter numbers, training token numbers, and token-level sparsity. We involve four parameter scales: Small (0.1B), Medium (0.5B), Large (0.8B), and XLarge (1.2B). See Appendix~\ref{sec:model-setting} for model settings.

\begin{table*}[t]
    \footnotesize
    \centering
    \setlength{\tabcolsep}{0.3em}
    \begin{tabular}{lccc|ccc|ccc|ccc}
    \toprule
    \multirow{2}{*}{Setting} & \multicolumn{3}{c|}{Small} & \multicolumn{3}{c|}{Medium} & \multicolumn{3}{c|}{Large} & \multicolumn{3}{c}{XLarge} \\
    \cmidrule{2-13}
    & $TLS$ & $CLS_8\uparrow$ & PPL$\downarrow$ & $TLS$ & $CLS_8\uparrow$ & PPL$\downarrow$ & $TLS$ & $CLS_8\uparrow$ & PPL$\downarrow$ & $TLS$ & $CLS_8\uparrow$ & PPL$\downarrow$ \\
    \midrule

    Dense & - & - & 14.90 & - & - & 10.03 & - & - & 9.29 & - & - & 8.49 \\
    \midrule

    TopK & 79.17 & 49.18 & 16.22 & 85.00 & 62.25 & 10.58 & 83.33 & 59.40 & 9.96 & 82.14 & 61.05 & 8.87 \\
    DSMoE  & 79.17 & 49.27 & 15.53 & 85.00 & 66.22 & 10.69 & 83.33 & 62.06 & 9.89 & 82.14 & 60.28 & 8.86 \\
    GRIN         & 79.17 & 50.45 & 15.50 & 85.00 & 61.48 & 10.40 & 83.33 & 59.08 & 9.72 & 82.14 & 60.89 & 9.03 \\
    ReMoE        & 78.33 & 42.44 & \textbf{14.60} & 84.43 & 52.00 & 10.42 & 82.80 & 50.79 & 9.60 & 81.93 & 51.01 & 8.78 \\

    \midrule

    \textbf{BlockFFN} & 80.54 & \textbf{71.38} & 14.88 & 84.25 & \textbf{75.87} & \textbf{10.23} & 84.05 & \textbf{73.79} & \textbf{9.52} & 81.81 & \textbf{72.78} & \textbf{8.69} \\
    \bottomrule
    \end{tabular}
    \caption{The average perplexities (PPL) and chunk-level sparsity for 8 consecutive tokens ($CLS_8$) on the validation data under close TLS. ``Dense'' is the upper bound setting, which involves vanilla Transformers with the same parameter numbers as MoE settings.}
    \label{tab:moe-ppl}
\end{table*}

\begin{table*}[t]
    \footnotesize
    \centering
    \begin{tabular}{lcc|cc|cc|cc}
    \toprule
    \multirow{2}{*}{Setting} & \multicolumn{2}{c|}{Small} & \multicolumn{2}{c|}{Medium} & \multicolumn{2}{c|}{Large} & \multicolumn{2}{c}{XLarge} \\
    \cmidrule{2-9}
    & C.R.$\uparrow$ & R.C.$\uparrow$ & C.R.$\uparrow$ & R.C.$\uparrow$ & C.R.$\uparrow$ & R.C.$\uparrow$ & C.R.$\uparrow$ & R.C.$\uparrow$ \\
    \midrule

    Dense & 45.15 & 32.72 & 52.68 & 44.85 & 54.89 & 48.87 & 57.30 & 49.53 \\
    \midrule

    TopK & 43.73 & 28.99 & 50.90 & 42.47 & 52.84 & 43.05 & 54.98 & 49.29 \\
    DSMoE  & 44.35 & 31.90 & 50.46 & \textbf{44.89} & 52.54 & 46.45 & 55.28 & 50.57 \\
    GRIN         & 43.69 & 31.61 & 50.96 & 42.56 & 52.93 & 45.04 & 54.65 & 49.68 \\
    ReMoE        & \textbf{45.22} & 32.78 & 51.42 & 43.05 & 53.77 & 47.33 & 55.56 & 47.12 \\

    \midrule

    \textbf{BlockFFN} & 44.80 & \textbf{33.93} & \textbf{51.75} & 43.74 & \textbf{54.44} &\textbf{51.60} & \textbf{56.42} & \textbf{50.73} \\
    \bottomrule
    \end{tabular}
    \caption{The average evaluation scores on two groups of benchmarks: commonsense reasoning (C.R.) and reading comprehension (R.C.). ``Dense'' is the upper bound setting.}
    \label{tab:moe-eval}
    \vspace{-1.5em}
\end{table*}

We adopt two comparison metrics: perplexity (PPL) on validation datasets and evaluation scores on benchmarks. Benchmarks include two groups: commonsense reasoning (C.R.) and reading comprehension (R.C.). See Appendix~\ref{sec:data-benchmark} for details about data and benchmarks.

The PPL and evaluation scores are shown in Table~\ref{tab:moe-ppl} and~\ref{tab:moe-eval}, respectively. The training curves of the "XLarge" settings are drawn in Figure~\ref{fig:train-curve-12b}. We can draw the following observations:

(1) \textit{Performance}: Under close parameter numbers, all the settings (except for Small ReMoE and BlockFFN) cannot match the ``Dense'' setting, due to the performance compromise of sparsification. However, under close TLS values (i.e., identical average FLOPs for each token), \textbf{BlockFFN outperforms other MoE baselines in terms of validation PPL, train loss, and scores on downstream tasks}, showing less performance compromise.

(2) \textit{Sparsity}: Under close TLS values, \textbf{BlockFFN always has considerably higher CLS values than other baselines}. Attributed to CLS-oriented training objectives, this property makes BlockFFN more friendly for acceleration.

\subsubsection{Expert Selection Stability}

Low-resource conditions often require the implementation of memory-saving techniques, such as offloading, where the weights of experts are loaded into memory only when they are activated. Such a technique calls for higher expert selection stability. Specifically, the distribution of selected experts should be as similar as possible across consecutive tokens, so that the costs of expert IO can be saved.

In this section, we demonstrate that BlockFFN has significant expert selection stability, which is measured by the \textbf{reuse ratio}, namely, within the activated experts of one token, the average ratio of experts that are also activated by its next token. Within a sequence with $L>1$ tokens, the set of experts activated by the $i$-th token is denoted by $\mathcal{S}_i$. The reuse ratio of this sequence is calculated by $\frac{1}{L-1}\sum_i^{L-1}\frac{|\mathcal{S}_i\cap\mathcal{S}_{i+1}|}{|\mathcal{S}_i|}$. As shown in Table~\ref{tab:reuse-ratio}, the high reuse ratios of BlockFFN models over 85\% ensure satisfactory memory efficiency and good adaptability to offloading.

\begin{table*}[ht]
    \vspace{-1em}
    \footnotesize
    \centering
    \begin{tabular}{l|cccc}
    \toprule
    Scale & Small & Medium & Large & XLarge \\
    \midrule
    Reuse Ratio (\%) & 90.28 & 89.69 & 87.13 & 89.57 \\
    \bottomrule
    \end{tabular}
    \caption{The average reuse ratios of BlockFFN models on the validation data.}
    \label{tab:reuse-ratio}
    \vspace{-1.5em}
\end{table*}

\subsubsection{Analysis of Expert Allocation}

\begin{figure}[t]
\begin{minipage}{0.48\textwidth}
    \centering
    \includegraphics[width=0.8\linewidth]{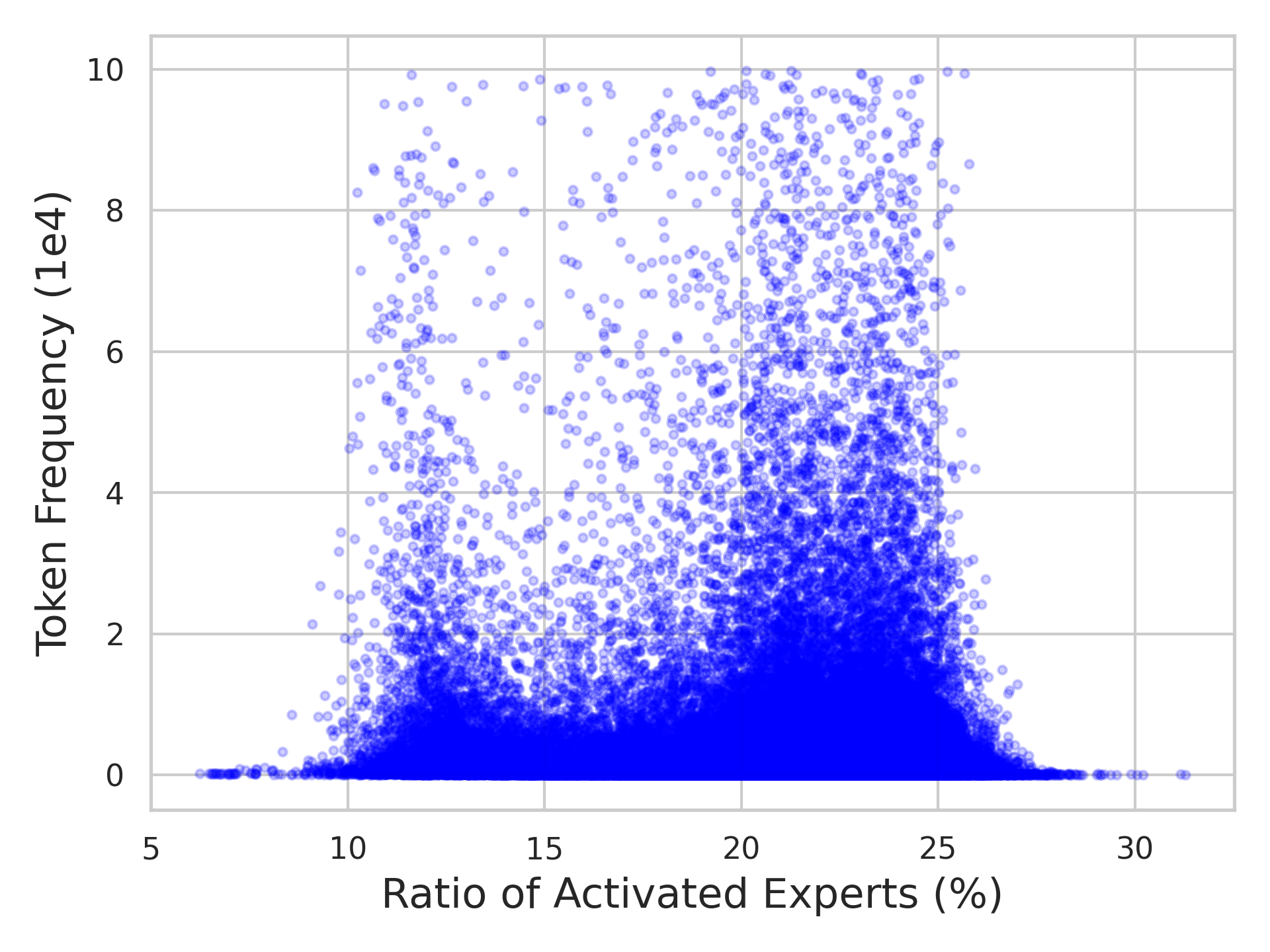}
    \caption{For each token in vocabulary, we calculate its frequencies and average ratios of activated experts, which show a bimodal distribution of expert allocation.}
    \label{fig:freq-act}
\end{minipage}
\hfill
\begin{minipage}{0.48\textwidth}
    \centering
    \includegraphics[width=0.8\linewidth]{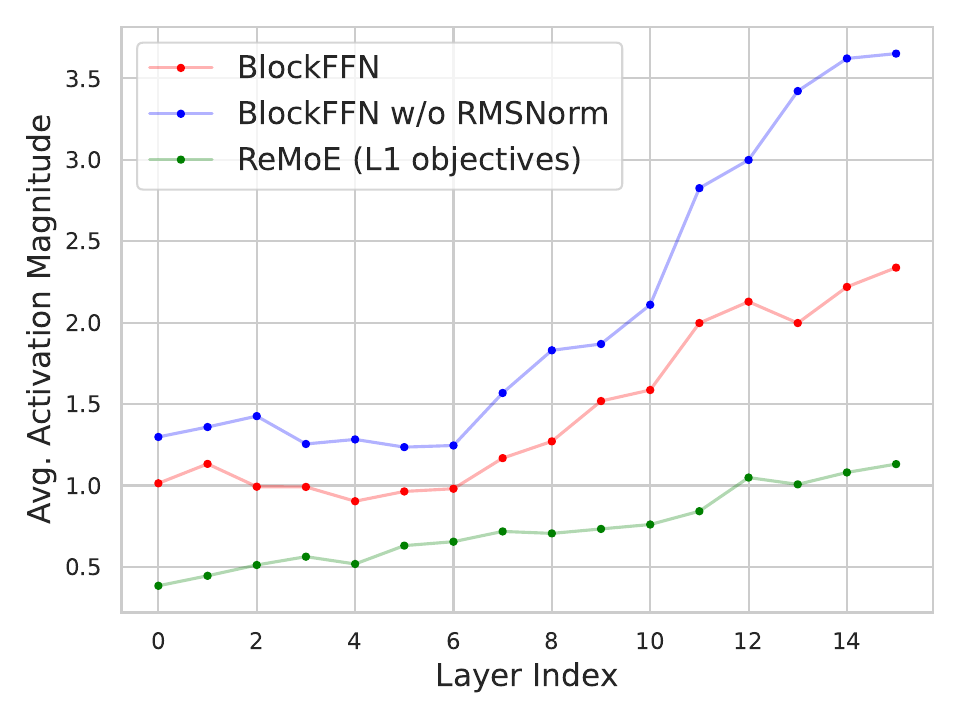}
    \caption{The layer-wise distributions of average activation magnitudes on the ``Small'' settings. While BlockFFN uses CLS-aware objectives, ReMoE adopts L1 regularization.}
    \label{fig:layer-magnitude}
\end{minipage}
\vspace{-1em}
\end{figure}

While ReLU-based routing is intrinsically differentiable, in this section, we examine how the router allocates experts and whether the inflexibility of activation patterns is truly addressed. Based on the validation data, we calculated the frequencies and average ratios of activated experts for each token, which are shown in Figure~\ref{fig:freq-act}. These results demonstrate a \textbf{bimodal distribution of expert allocation}.

Specifically, the smaller peak lies between the activation ratio interval between 10\% and 15\%, which involves tokens such as numbers (e.g., ``\texttt{0}'', ``\texttt{1}''), single characters (e.g., ``\texttt{a}'', ``\texttt{b}''), and reserved words of programs (e.g., ``\texttt{import}'', ``\texttt{return}''). These tokens have more deterministic meanings and thus require fewer experts for processing. By contrast, the larger peak between 20\% and 25\% mainly involves those tokens with more complex or diverse meanings, such as English pronouns and Chinese characters. Therefore, they need more experts to understand. Such a bimodal allocation of experts demonstrates that ReLU activation can truly address the routing inflexibility and allocate resources more wisely.

\subsubsection{RMSNorm and Activation Magnitude Disturbance} \label{sec:rmsnorm-disturbance}

In this section, we examine the effectiveness of the RMSNorm in our router module. First, we conduct an ablation study on the ``Small'' setting. \textbf{After removing the RMSNorm layer, the validation PPL rises from 14.88 to 15.04, indicating the effectiveness of RMSNorm.}

Moreover, to inspect the effects of RMSNorm, we calculate the average activation magnitudes (computed by L2 Norm) on the validation data. As shown in Figure~\ref{fig:layer-magnitude}, under all settings, higher layers (closer to output) generally have larger activation magnitudes. Without RMSNorm, the magnitudes of activation values in BlockFFN considerably rise with worse performance. By contrast, ReMoE, which has a similar architecture to BlockFFN without RMSNorm, suffers from significantly smaller activation magnitudes. This is attributed to the activation shrinkage issue induced by L1 regularization directly imposed on activation values~\citep{rajamanoharan2024improving}. We assume that RMSNorm, preventing activation values from direct regularization (Section~\ref{sec:blockffn-arch}), potentially alleviates activation magnitude disturbance and maintains a more stable and appropriate magnitude level.

Besides the above issues, expert granularity also has an important influence. Through experiments, we find that the validation loss generally decreases with finer experts, but the marginal benefits quickly diminish with $>40$ experts for BlockFFN Medium. However, \textbf{the relationship between sparsity and expert granularity is nonmonotonic}, with the best setting of 40 experts achieving the highest sparsity. See Appendix~\ref{sec:expert-granularity} for more details.

\begin{table*}[t]
    \footnotesize
    \centering
    \begin{tabular}{lccc|lccc}
    \toprule
    \multicolumn{4}{c|}{Direct Ablation} & \multicolumn{4}{c}{Substitute Objectives} \\
    \midrule
    Setting & $TLS$ & $CLS_8\uparrow$ & PPL$\downarrow$ & Setting & $TLS$ & $CLS_8\uparrow$ & PPL$\downarrow$ \\
    \midrule
    \underline{\textbf{AL+CS}} & 80.54 & 71.38 & 14.88 & AL+L1 & 79.86 & 65.16 & 15.50 \\
    CS    & 81.67 & 67.56 & 15.66 & AL+Ent & 81.05 & 69.13 & 15.97 \\
    AL    & 63.55 & 52.59 & 14.89 & L1  & 79.35 & 45.13 & 15.01 \\
    Null  & 48.56 & 14.89 & 14.85 & Ent & 79.04 & 45.78 & 15.12 \\
    \bottomrule
    \end{tabular}
    \caption{Ablation studies on the training objectives. ``\textbf{AL+CS}'' is our standard setting. ``AL'', ``CS'', ``L1'' and ``Ent'' indicates activation locality, chunk sparsification, L1 norm, and router entropy, respectively. ``Null'' is the setting without any additional training objectives.}
    \label{tab:loss-ablation}
    \vspace{-1em}
\end{table*}

\subsection{Training Objective Rationality}

In Section~\ref{sec:training-objective}, we introduce \textbf{activation locality (AL) loss} and \textbf{chunk sparsification (CS) loss} as our training objectives. In this part, we mainly discuss whether such a practice is reasonable and better than other potential substitutes.

First, we conduct direct ablation studies by removing either AL loss or CS loss. As shown in the left part of Table~\ref{tab:loss-ablation}, without AL (Setting ``CS''), the model suffers from lower CLS and considerably higher PPL. On the other hand, without ``CS'' (Setting ``AL'' and ``Null''), the sparsity (both $TLS$ and $CLS$) can be extremely low. These demonstrate the division of labor: \textbf{CS is mainly responsible for global sparsification, while AL is to promote the CLS with less performance compromise (compared with the direct application of a large CS loss)}. A possible explanation for why ``AL+CS'' performs better than pure ``CS'' is the competing relationship between ``AL'' and ``CS''. Specifically, the introduction of ``AL'' weakens the sparsification effect of ``CS'', producing lower TLS but higher CLS, and the better performance is attributed to the lower TLS level.

Next, we explore other potential substitute training objectives for sparsification. This includes the L1 norm (L1)~\citep{song2024prosparse} and router entropy loss (Ent)~\citep{huang2024harder}. As shown in the right part of Table~\ref{tab:loss-ablation}, replacing CS with L1/Ent (Setting ``AL+L1'' and ``AL+Ent'') can cause a considerable drop in performance. Besides, due to the absence of AL, ``L1'' and ``Ent'' cannot reach satisfactory CLS, either. Therefore, \textbf{CS is a more competitive sparsification partner of AL with less performance compromise}.

\begin{table*}[t]
    \footnotesize
    \centering
    \begin{tabular}{lcccccc|c}
    \toprule
    Setting & \textbf{MT.} & \textbf{Trans.} & \textbf{Summ.} & \textbf{QA} & \textbf{Math} & \textbf{RAG} & \textbf{Average} \\
    \midrule
    Huggingface & 7.61 & 7.32 & 7.54 & 7.81 & 7.76 & 5.96 & 7.33 (0.57$\times$) \\
    \midrule
    Baseline AR & 12.61 & 13.32 & 12.40 & 13.04 & 12.80 & 12.86 & 12.84 (1.00$\times$) \\
    EAGLE-2 & 23.70 & 20.26 & 20.74 & 21.99 & 25.01 & 22.47 & 22.36 (1.74$\times$) \\
    \midrule
    Ours (1-Tok) & 38.41 & \textbf{40.39} & 37.37 & 42.05 & 44.54 & 39.38 & 40.36 (3.14$\times$) \\
    \textbf{Ours (32-Tok)} & \textbf{49.43} & 39.18 & \textbf{42.70} & \textbf{46.09} & \textbf{59.85} & \textbf{45.76} & \textbf{47.17} (\textbf{3.67}$\times$) \\
    \bottomrule
    \end{tabular}
    \caption{Decoding speeds (token/sec) and average speedup ratios on NVIDIA Jetson Orin NX. ``Ours (1-Tok)'' is our token-level acceleration kernel purely dependent on sparsity, while ``Ours (32-Tok)'' is our efficient chunk-level acceleration kernels that combine EAGLE-2 and chunk-level sparsity. The speedup ratios are relative to ``Baseline AR''.}
    \label{tab:speedup-overall}
    \vspace{-1.5em}
\end{table*}

\subsection{Practical Inference Acceleration}

\textbf{Speedup experiment}\quad To demonstrate the efficacy of our acceleration kernels, we conduct speedup experiments on Spec-Bench~\citep{xia2024unlocking}, a comprehensive benchmark for speculative decoding, with NVIDIA Jetson Orin NX 16GB. To ensure comparison fairness, except for the vanilla Huggingface auto-regressive decoding, all baseline methods are implemented within the framework of FR-Spec~\citep{zhao2025fr}, which applies CUDA kernels to reduce IO overheads and is much faster than Huggingface. These baselines include Baseline AR (i.e., a faster FR-Spec auto-regressive implementation), and EAGLE-2~\citep{li2024eagle2}. Moreover, since acceleration effects are more significant on larger models,  we specifically train a 2.8B BlockFFN model as the base of our efficiency experiment.

From Table~\ref{tab:speedup-overall}, we have the following observations:
(1) ``Baseline AR'' is considerably faster than ``Huggingface'', indicating that our experimental framework is efficient enough and can alleviate the influence of potential IO overheads.
(2) ``Ours (32-Tok)'', our acceleration kernel combining speculative decoding and activation sparsity, achieves the highest decoding speed, with 3.67$\times$ speedup. Meanwhile, it is faster than the pure sparsity setting ``Ours (1-Tok)'' and the pure speculative decoding ``EAGLE-2''. This demonstrates the value of such a combination and reveals the significant value of utilizing activation sparsity in end-side device inference acceleration. See Appendix~\ref{sec:speedup-independent} for speedup data on independent datasets.

\textbf{Upper bound analysis}\quad Moreover, we conduct further experiments and find that both ``Ours (1-Tok)'' and ``Ours (32-Tok)'' can reach the theoretical upper bound of FFN speedup ratios induced by the corresponding token-level sparsity and the average union sparsity of tokens contained in an EAGLE-2 draft tree, see Appendix~\ref{sec:upper-bound}.

\section{Conclusion}

In this work, we propose BlockFFN, a novel MoE architecture equipped with a ReLU-based differentiable and flexible routing strategy, which enables BlockFFN to outperform existing MoE counterparts. Next, we advocate more attention to chunk-level sparsity (CLS), and introduce the CLS-aware training objectives to promote the 8-token CLS to over 70\%, offering BlockFFN better activation locality and more friendliness for end-side device acceleration. Finally, our efficient acceleration kernels achieve up to 3.67$\times$ speedup on NVIDIA Jetson Orin NX than the baseline auto-regressive decoding, reaching the sparsity-induced upper bound of FFN acceleration.


\section*{Acknowledgments}


This work is supported by the National Key R\&D Program of China (No.2022ZD0116312), Beijing Municipal Science and Technology Plan Project
(Z241100001324025) and a grant from the Guoqiang Institute, Tsinghua University. Our research is also supported by Huawei and can be carried out using the Huawei Ascend AI technology stack.


\bibliography{colm2025_conference}

\begin{thebibliography}{64}
\providecommand{\natexlab}[1]{#1}
\providecommand{\url}[1]{\texttt{#1}}
\expandafter\ifx\csname urlstyle\endcsname\relax
  \providecommand{\doi}[1]{doi: #1}\else
  \providecommand{\doi}{doi: \begingroup \urlstyle{rm}\Url}\fi

\bibitem[Bisk et~al.(2020)Bisk, Zellers, Gao, Choi, et~al.]{piqa}
Yonatan Bisk, Rowan Zellers, Jianfeng Gao, Yejin Choi, et~al.
\newblock {PIQA}: Reasoning about physical commonsense in natural language.
\newblock In \emph{Proceedings of the AAAI conference on artificial intelligence}, volume~34, pp.\  7432--7439, 2020.
\newblock URL \url{https://ojs.aaai.org/index.php/AAAI/article/view/6239/6095}.

\bibitem[Cai et~al.(2024)Cai, Li, Geng, Peng, Lee, Chen, and Dao]{cai2024medusa}
Tianle Cai, Yuhong Li, Zhengyang Geng, Hongwu Peng, Jason~D Lee, Deming Chen, and Tri Dao.
\newblock Medusa: Simple {LLM} inference acceleration framework with multiple decoding heads.
\newblock In \emph{Forty-first International Conference on Machine Learning}, 2024.
\newblock URL \url{https://arxiv.org/pdf/2401.10774}.

\bibitem[Clark et~al.(2019)Clark, Lee, Chang, Kwiatkowski, Collins, and Toutanova]{boolq}
Christopher Clark, Kenton Lee, Ming-Wei Chang, Tom Kwiatkowski, Michael Collins, and Kristina Toutanova.
\newblock {BoolQ}: Exploring the surprising difficulty of natural yes/no questions.
\newblock In \emph{Proceedings of the 2019 Conference of the North American Chapter of the Association for Computational Linguistics: Human Language Technologies, Volume 1 (Long and Short Papers)}, pp.\  2924--2936, 2019.
\newblock URL \url{https://aclanthology.org/N19-1300.pdf}.

\bibitem[Clark et~al.(2020)Clark, Choi, Collins, Garrette, Kwiatkowski, Nikolaev, and Palomaki]{tydiqa}
Jonathan~H Clark, Eunsol Choi, Michael Collins, Dan Garrette, Tom Kwiatkowski, Vitaly Nikolaev, and Jennimaria Palomaki.
\newblock {TyDi QA}: A benchmark for information-seeking question answering in typologically diverse languages.
\newblock \emph{Transactions of the Association for Computational Linguistics}, 8:\penalty0 454--470, 2020.
\newblock URL \url{https://aclanthology.org/2020.tacl-1.30.pdf}.

\bibitem[Dai et~al.(2024)Dai, Deng, Zhao, Xu, Gao, Chen, Li, Zeng, Yu, Wu, et~al.]{dai2024deepseekmoe}
Damai Dai, Chengqi Deng, Chenggang Zhao, RX~Xu, Huazuo Gao, Deli Chen, Jiashi Li, Wangding Zeng, Xingkai Yu, Y~Wu, et~al.
\newblock {DeepSeekMoE}: Towards ultimate expert specialization in mixture-of-experts language models.
\newblock \emph{CoRR}, 2024.
\newblock URL \url{http://arxiv.org/pdf/2401.06066}.

\bibitem[Dauphin et~al.(2017)Dauphin, Fan, Auli, and Grangier]{dauphin2017language}
Yann~N Dauphin, Angela Fan, Michael Auli, and David Grangier.
\newblock Language modeling with gated convolutional networks.
\newblock In \emph{International Conference on Machine Learning}, pp.\  933--941. PMLR, 2017.
\newblock URL \url{https://proceedings.mlr.press/v70/dauphin17a/dauphin17a.pdf}.

\bibitem[Ding et~al.(2023)Ding, Chen, Xu, Qin, Zheng, Hu, Liu, Sun, and Zhou]{ding2023enhancing}
Ning Ding, Yulin Chen, Bokai Xu, Yujia Qin, Zhi Zheng, Shengding Hu, Zhiyuan Liu, Maosong Sun, and Bowen Zhou.
\newblock Enhancing chat language models by scaling high-quality instructional conversations.
\newblock \emph{arXiv preprint arXiv:2305.14233}, 2023.
\newblock URL \url{https://arxiv.org/pdf/2305.14233.pdf}.

\bibitem[Dubey et~al.(2024)Dubey, Jauhri, Pandey, Kadian, Al-Dahle, Letman, Mathur, Schelten, Yang, Fan, et~al.]{dubey2024llama}
Abhimanyu Dubey, Abhinav Jauhri, Abhinav Pandey, Abhishek Kadian, Ahmad Al-Dahle, Aiesha Letman, Akhil Mathur, Alan Schelten, Amy Yang, Angela Fan, et~al.
\newblock The {Llama} 3 herd of models.
\newblock \emph{arXiv preprint arXiv:2407.21783}, 2024.
\newblock URL \url{https://arxiv.org/pdf/2407.21783}.

\bibitem[Fedus et~al.(2022)Fedus, Zoph, and Shazeer]{fedus2022switch}
William Fedus, Barret Zoph, and Noam Shazeer.
\newblock Switch {Transformers}: Scaling to trillion parameter models with simple and efficient sparsity.
\newblock \emph{Journal of Machine Learning Research}, 23\penalty0 (120):\penalty0 1--39, 2022.
\newblock URL \url{https://www.jmlr.org/papers/volume23/21-0998/21-0998.pdf}.

\bibitem[Frantar \& Alistarh(2023)Frantar and Alistarh]{frantar2023sparsegpt}
Elias Frantar and Dan Alistarh.
\newblock {SparseGPT}: Massive language models can be accurately pruned in one-shot.
\newblock In \emph{International Conference on Machine Learning}, pp.\  10323--10337. PMLR, 2023.
\newblock URL \url{https://proceedings.mlr.press/v202/frantar23a/frantar23a.pdf}.

\bibitem[Gale et~al.(2023)Gale, Narayanan, Young, and Zaharia]{gale2023megablocks}
Trevor Gale, Deepak Narayanan, Cliff Young, and Matei Zaharia.
\newblock {MegaBlocks}: Efficient sparse training with mixture-of-experts.
\newblock \emph{Proceedings of Machine Learning and Systems}, 5:\penalty0 288--304, 2023.
\newblock URL \url{https://proceedings.mlsys.org/paper_files/paper/2023/file/5a54f79333768effe7e8927bcccffe40-Paper-mlsys2023.pdf}.

\bibitem[Gao et~al.(2020)Gao, Biderman, Black, Golding, Hoppe, Foster, Phang, He, Thite, Nabeshima, et~al.]{gao2020pile}
Leo Gao, Stella Biderman, Sid Black, Laurence Golding, Travis Hoppe, Charles Foster, Jason Phang, Horace He, Anish Thite, Noa Nabeshima, et~al.
\newblock The {P}ile: An 800{GB} dataset of diverse text for language modeling.
\newblock \emph{arXiv preprint arXiv:2101.00027}, 2020.
\newblock URL \url{https://arxiv.org/pdf/2101.00027.pdf}.

\bibitem[Gu et~al.(2023)Gu, Dong, Wei, and Huang]{gu2023knowledge}
Yuxian Gu, Li~Dong, Furu Wei, and Minlie Huang.
\newblock Knowledge distillation of large language models.
\newblock \emph{arXiv preprint arXiv:2306.08543}, 2023.
\newblock URL \url{https://arxiv.org/pdf/2306.08543.pdf}.

\bibitem[He et~al.(2021)He, Qiu, Zeng, Yang, Zhai, and Tang]{he2021fastmoe}
Jiaao He, Jiezhong Qiu, Aohan Zeng, Zhilin Yang, Jidong Zhai, and Jie Tang.
\newblock {FastMoE}: A fast mixture-of-expert training system.
\newblock \emph{arXiv preprint arXiv:2103.13262}, 2021.
\newblock URL \url{https://arxiv.org/pdf/2103.13262}.

\bibitem[Hsieh et~al.(2023)Hsieh, Li, Yeh, Nakhost, Fujii, Ratner, Krishna, Lee, and Pfister]{hsieh2023distilling}
Cheng-Yu Hsieh, Chun-Liang Li, Chih-Kuan Yeh, Hootan Nakhost, Yasuhisa Fujii, Alexander Ratner, Ranjay Krishna, Chen-Yu Lee, and Tomas Pfister.
\newblock Distilling step-by-step! outperforming larger language models with less training data and smaller model sizes.
\newblock \emph{arXiv preprint arXiv:2305.02301}, 2023.
\newblock URL \url{https://arxiv.org/pdf/2305.02301.pdf}.

\bibitem[Hu et~al.(2024)Hu, Tu, Han, He, Cui, Long, Zheng, Fang, Huang, Zhao, et~al.]{hu2024minicpm}
Shengding Hu, Yuge Tu, Xu~Han, Chaoqun He, Ganqu Cui, Xiang Long, Zhi Zheng, Yewei Fang, Yuxiang Huang, Weilin Zhao, et~al.
\newblock {MiniCPM}: Unveiling the potential of small language models with scalable training strategies.
\newblock \emph{arXiv preprint arXiv:2404.06395}, 2024.

\bibitem[Huang et~al.(2024)Huang, An, Zhuang, Tao, Zhang, Jin, Xu, Chen, Huang, and Feng]{huang2024harder}
Quzhe Huang, Zhenwei An, Nan Zhuang, Mingxu Tao, Chen Zhang, Yang Jin, Kun Xu, Liwei Chen, Songfang Huang, and Yansong Feng.
\newblock Harder tasks need more experts: Dynamic routing in {MoE} models.
\newblock \emph{arXiv preprint arXiv:2403.07652}, 2024.
\newblock URL \url{https://arxiv.org/pdf/2403.07652}.

\bibitem[Hwang et~al.(2023)Hwang, Cui, Xiong, Yang, Liu, Hu, Wang, Salas, Jose, Ram, et~al.]{hwang2023tutel}
Changho Hwang, Wei Cui, Yifan Xiong, Ziyue Yang, Ze~Liu, Han Hu, Zilong Wang, Rafael Salas, Jithin Jose, Prabhat Ram, et~al.
\newblock Tutel: Adaptive mixture-of-experts at scale.
\newblock \emph{Proceedings of Machine Learning and Systems}, 5:\penalty0 269--287, 2023.
\newblock URL \url{https://proceedings.mlsys.org/paper_files/paper/2023/file/5616d34cf8ff73942cfd5aa922842556-Paper-mlsys2023.pdf}.

\bibitem[Jiang et~al.(2024)Jiang, Sablayrolles, Roux, Mensch, Savary, Bamford, Chaplot, Casas, Hanna, Bressand, et~al.]{jiang2024mixtral}
Albert~Q Jiang, Alexandre Sablayrolles, Antoine Roux, Arthur Mensch, Blanche Savary, Chris Bamford, Devendra~Singh Chaplot, Diego de~las Casas, Emma~Bou Hanna, Florian Bressand, et~al.
\newblock Mixtral of experts.
\newblock \emph{arXiv preprint arXiv:2401.04088}, 2024.
\newblock URL \url{https://arxiv.org/pdf/2401.04088}.

\bibitem[Krajewski et~al.(2024)Krajewski, Ludziejewski, Adamczewski, Pi{\'o}ro, Krutul, Antoniak, Ciebiera, Kr{\'o}l, Odrzyg{\'o}{\'z}d{\'z}, Sankowski, et~al.]{krajewski2024scaling}
Jakub Krajewski, Jan Ludziejewski, Kamil Adamczewski, Maciej Pi{\'o}ro, Micha{\l} Krutul, Szymon Antoniak, Kamil Ciebiera, Krystian Kr{\'o}l, Tomasz Odrzyg{\'o}{\'z}d{\'z}, Piotr Sankowski, et~al.
\newblock Scaling laws for fine-grained mixture of experts.
\newblock \emph{arXiv preprint arXiv:2402.07871}, 2024.
\newblock URL \url{https://arxiv.org/pdf/2402.07871}.

\bibitem[Leviathan et~al.(2023)Leviathan, Kalman, and Matias]{leviathan2023fast}
Yaniv Leviathan, Matan Kalman, and Yossi Matias.
\newblock Fast inference from {Transformers} via speculative decoding.
\newblock In \emph{International Conference on Machine Learning}, pp.\  19274--19286. PMLR, 2023.
\newblock URL \url{https://proceedings.mlr.press/v202/leviathan23a/leviathan23a.pdf}.

\bibitem[Li et~al.(2023)Li, Allal, Zi, Muennighoff, Kocetkov, Mou, Marone, Akiki, Li, Chim, et~al.]{li2023starcoder}
Raymond Li, Loubna~Ben Allal, Yangtian Zi, Niklas Muennighoff, Denis Kocetkov, Chenghao Mou, Marc Marone, Christopher Akiki, Jia Li, Jenny Chim, et~al.
\newblock {StarCoder}: may the source be with you!
\newblock \emph{arXiv preprint arXiv:2305.06161}, 2023.
\newblock URL \url{https://arxiv.org/pdf/2305.06161.pdf}.

\bibitem[Li et~al.(2024{\natexlab{a}})Li, Wei, Zhang, and Zhang]{li2024eagle}
Yuhui Li, Fangyun Wei, Chao Zhang, and Hongyang Zhang.
\newblock {EAGLE}: Speculative sampling requires rethinking feature uncertainty.
\newblock In \emph{Forty-first International Conference on Machine Learning}, 2024{\natexlab{a}}.
\newblock URL \url{https://arxiv.org/pdf/2401.15077}.

\bibitem[Li et~al.(2024{\natexlab{b}})Li, Wei, Zhang, and Zhang]{li2024eagle2}
Yuhui Li, Fangyun Wei, Chao Zhang, and Hongyang Zhang.
\newblock {EAGLE-2}: Faster inference of language models with dynamic draft trees.
\newblock In \emph{Proceedings of the 2024 Conference on Empirical Methods in Natural Language Processing}, pp.\  7421--7432, 2024{\natexlab{b}}.
\newblock URL \url{https://aclanthology.org/2024.emnlp-main.422.pdf}.

\bibitem[Li et~al.(2022)Li, You, Bhojanapalli, Li, Rawat, Reddi, Ye, Chern, Yu, Guo, et~al.]{li2022lazy}
Zonglin Li, Chong You, Srinadh Bhojanapalli, Daliang Li, Ankit~Singh Rawat, Sashank~J Reddi, Ke~Ye, Felix Chern, Felix Yu, Ruiqi Guo, et~al.
\newblock The lazy neuron phenomenon: On emergence of activation sparsity in {Transformers}.
\newblock In \emph{The Eleventh International Conference on Learning Representations}, 2022.
\newblock URL \url{https://openreview.net/pdf?id=TJ2nxciYCk-}.

\bibitem[Liu et~al.(2024{\natexlab{a}})Liu, Feng, Wang, Wang, Liu, Zhao, Dengr, Ruan, Dai, Guo, et~al.]{liu2024deepseekv2}
Aixin Liu, Bei Feng, Bin Wang, Bingxuan Wang, Bo~Liu, Chenggang Zhao, Chengqi Dengr, Chong Ruan, Damai Dai, Daya Guo, et~al.
\newblock {DeepSeek-V2}: A strong, economical, and efficient mixture-of-experts language model.
\newblock \emph{arXiv preprint arXiv:2405.04434}, 2024{\natexlab{a}}.
\newblock URL \url{https://arxiv.org/pdf/2405.04434}.

\bibitem[Liu et~al.(2024{\natexlab{b}})Liu, Feng, Xue, Wang, Wu, Lu, Zhao, Deng, Zhang, Ruan, et~al.]{liu2024deepseek}
Aixin Liu, Bei Feng, Bing Xue, Bingxuan Wang, Bochao Wu, Chengda Lu, Chenggang Zhao, Chengqi Deng, Chenyu Zhang, Chong Ruan, et~al.
\newblock {DeepSeek-V3} technical report.
\newblock \emph{arXiv preprint arXiv:2412.19437}, 2024{\natexlab{b}}.
\newblock URL \url{https://arxiv.org/pdf/2412.19437}.

\bibitem[Liu et~al.(2024{\natexlab{c}})Liu, Kim, Wang, Liang, Shen, Cheng, Liu, Tanaka, Wu, Hu, et~al.]{liu2024grin}
Liyuan Liu, Young~Jin Kim, Shuohang Wang, Chen Liang, Yelong Shen, Hao Cheng, Xiaodong Liu, Masahiro Tanaka, Xiaoxia Wu, Wenxiang Hu, et~al.
\newblock {GRIN}: Gradient-informed {MoE}.
\newblock \emph{arXiv preprint arXiv:2409.12136}, 2024{\natexlab{c}}.
\newblock URL \url{https://arxiv.org/pdf/2409.12136}.

\bibitem[Liu et~al.(2023)Liu, Wang, Dao, Zhou, Yuan, Song, Shrivastava, Zhang, Tian, Re, et~al.]{liu2023deja}
Zichang Liu, Jue Wang, Tri Dao, Tianyi Zhou, Binhang Yuan, Zhao Song, Anshumali Shrivastava, Ce~Zhang, Yuandong Tian, Christopher Re, et~al.
\newblock {Deja Vu}: Contextual sparsity for efficient {LLMs} at inference time.
\newblock In \emph{International Conference on Machine Learning}, pp.\  22137--22176. PMLR, 2023.
\newblock URL \url{https://proceedings.mlr.press/v202/liu23am/liu23am.pdf}.

\bibitem[Luo et~al.(2024)Luo, Song, Han, Chen, Xiao, Liu, and Sun]{luo2024sparsing}
Yuqi Luo, Chenyang Song, Xu~Han, Yingfa Chen, Chaojun Xiao, Zhiyuan Liu, and Maosong Sun.
\newblock {S}parsing {L}aw: Towards large language models with greater activation sparsity.
\newblock \emph{arXiv preprint arXiv:2411.02335}, 2024.
\newblock URL \url{https://arxiv.org/pdf/2411.02335}.

\bibitem[Ma et~al.(2023)Ma, Fang, and Wang]{ma2023llm}
Xinyin Ma, Gongfan Fang, and Xinchao Wang.
\newblock {LLM-Pruner}: On the structural pruning of large language models.
\newblock \emph{arXiv preprint arXiv:2305.11627}, 2023.
\newblock URL \url{https://arxiv.org/pdf/2305.11627.pdf}.

\bibitem[Mirzadeh et~al.(2023)Mirzadeh, Alizadeh, Mehta, Del~Mundo, Tuzel, Samei, Rastegari, and Farajtabar]{mirzadeh2023relu}
Iman Mirzadeh, Keivan Alizadeh, Sachin Mehta, Carlo~C Del~Mundo, Oncel Tuzel, Golnoosh Samei, Mohammad Rastegari, and Mehrdad Farajtabar.
\newblock {ReLU} strikes back: Exploiting activation sparsity in large language models.
\newblock \emph{arXiv preprint arXiv:2310.04564}, 2023.
\newblock URL \url{https://arxiv.org/pdf/2310.04564.pdf}.

\bibitem[Muqeeth et~al.(2023)Muqeeth, Liu, and Raffel]{muqeeth2023soft}
Mohammed Muqeeth, Haokun Liu, and Colin Raffel.
\newblock Soft merging of experts with adaptive routing.
\newblock \emph{arXiv preprint arXiv:2306.03745}, 2023.
\newblock URL \url{https://arxiv.org/pdf/2306.03745}.

\bibitem[Paperno et~al.(2016)Paperno, Kruszewski, Lazaridou, Pham, Bernardi, Pezzelle, Baroni, Boleda, and Fern{\'a}ndez]{lambada}
Denis Paperno, Germ{\'a}n Kruszewski, Angeliki Lazaridou, Ngoc-Quan Pham, Raffaella Bernardi, Sandro Pezzelle, Marco Baroni, Gemma Boleda, and Raquel Fern{\'a}ndez.
\newblock The {LAMBADA} dataset: Word prediction requiring a broad discourse context.
\newblock In \emph{Proceedings of the 54th Annual Meeting of the Association for Computational Linguistics (Volume 1: Long Papers)}, pp.\  1525--1534, 2016.
\newblock URL \url{https://aclanthology.org/P16-1144.pdf}.

\bibitem[Raffel et~al.(2020)Raffel, Shazeer, Roberts, Lee, Narang, Matena, Zhou, Li, and Liu]{raffel2020exploring}
Colin Raffel, Noam Shazeer, Adam Roberts, Katherine Lee, Sharan Narang, Michael Matena, Yanqi Zhou, Wei Li, and Peter~J Liu.
\newblock Exploring the limits of transfer learning with a unified text-to-text {Transformer}.
\newblock \emph{Journal of machine learning research}, 21\penalty0 (140):\penalty0 1--67, 2020.
\newblock URL \url{https://www.jmlr.org/papers/volume21/20-074/20-074.pdf}.

\bibitem[Rajamanoharan et~al.(2024)Rajamanoharan, Conmy, Smith, Lieberum, Varma, Kram{\'a}r, Shah, and Nanda]{rajamanoharan2024improving}
Senthooran Rajamanoharan, Arthur Conmy, Lewis Smith, Tom Lieberum, Vikrant Varma, J{\'a}nos Kram{\'a}r, Rohin Shah, and Neel Nanda.
\newblock Improving dictionary learning with gated sparse autoencoders.
\newblock \emph{arXiv preprint arXiv:2404.16014}, 2024.
\newblock URL \url{https://arxiv.org/pdf/2404.16014}.

\bibitem[Ramachandran et~al.(2017)Ramachandran, Zoph, and Le]{ramachandran2017searching}
Prajit Ramachandran, Barret Zoph, and Quoc~V Le.
\newblock Searching for activation functions.
\newblock \emph{arXiv preprint arXiv:1710.05941}, 2017.
\newblock URL \url{https://arxiv.org/pdf/1710.05941}.

\bibitem[Sap et~al.(2019)Sap, Rashkin, Chen, Le~Bras, and Choi]{siqa}
Maarten Sap, Hannah Rashkin, Derek Chen, Ronan Le~Bras, and Yejin Choi.
\newblock {SocialIQA}: Commonsense reasoning about social interactions.
\newblock In \emph{Proceedings of the 2019 Conference on Empirical Methods in Natural Language Processing and the 9th International Joint Conference on Natural Language Processing (EMNLP-IJCNLP)}, pp.\  4463--4473, 2019.
\newblock URL \url{https://aclanthology.org/D19-1454.pdf}.

\bibitem[Shao et~al.(2023)Shao, Chen, Zhang, Xu, Zhao, Li, Zhang, Gao, Qiao, and Luo]{shao2023omniquant}
Wenqi Shao, Mengzhao Chen, Zhaoyang Zhang, Peng Xu, Lirui Zhao, Zhiqian Li, Kaipeng Zhang, Peng Gao, Yu~Qiao, and Ping Luo.
\newblock Omniquant: Omnidirectionally calibrated quantization for large language models.
\newblock In \emph{The Twelfth International Conference on Learning Representations}, 2023.
\newblock URL \url{https://arxiv.org/pdf/2308.13137}.

\bibitem[Shazeer(2020)]{shazeer2020glu}
Noam Shazeer.
\newblock {GLU} variants improve {Transformer}.
\newblock \emph{arXiv preprint arXiv:2002.05202}, 2020.
\newblock URL \url{https://arxiv.org/pdf/2002.05202.pdf}.

\bibitem[Soldaini et~al.(2024)Soldaini, Kinney, Bhagia, Schwenk, Atkinson, Authur, Bogin, Chandu, Dumas, Elazar, et~al.]{soldaini2024dolma}
Luca Soldaini, Rodney Kinney, Akshita Bhagia, Dustin Schwenk, David Atkinson, Russell Authur, Ben Bogin, Khyathi Chandu, Jennifer Dumas, Yanai Elazar, et~al.
\newblock Dolma: An open corpus of three trillion tokens for language model pretraining research.
\newblock \emph{arXiv preprint arXiv:2402.00159}, 2024.
\newblock URL \url{https://arxiv.org/pdf/2402.00159}.

\bibitem[Song et~al.(2025)Song, Han, Zhang, Hu, Shi, Li, Chen, Liu, Li, Yang, and Sun]{song2024prosparse}
Chenyang Song, Xu~Han, Zhengyan Zhang, Shengding Hu, Xiyu Shi, Kuai Li, Chen Chen, Zhiyuan Liu, Guangli Li, Tao Yang, and Maosong Sun.
\newblock {P}ro{S}parse: Introducing and enhancing intrinsic activation sparsity within large language models.
\newblock In \emph{Proceedings of the 31st International Conference on Computational Linguistics}, pp.\  2626--2644, January 2025.
\newblock URL \url{https://aclanthology.org/2025.coling-main.180.pdf}.

\bibitem[Song et~al.(2023)Song, Mi, Xie, and Chen]{song2023powerinfer}
Yixin Song, Zeyu Mi, Haotong Xie, and Haibo Chen.
\newblock {PowerInfer}: Fast large language model serving with a consumer-grade {GPU}.
\newblock \emph{arXiv preprint arXiv:2312.12456}, 2023.
\newblock URL \url{https://arxiv.org/pdf/2312.12456.pdf}.

\bibitem[Song et~al.(2024)Song, Xie, Zhang, Wen, Ma, Mi, and Chen]{song2024turbo}
Yixin Song, Haotong Xie, Zhengyan Zhang, Bo~Wen, Li~Ma, Zeyu Mi, and Haibo Chen.
\newblock Turbo {Sparse}: Achieving {LLM} {SOTA} performance with minimal activated parameters.
\newblock \emph{arXiv preprint arXiv:2406.05955}, 2024.
\newblock URL \url{https://arxiv.org/pdf/2406.05955}.

\bibitem[Sun et~al.(2023)Sun, Liu, Bair, and Kolter]{sun2023simple}
Mingjie Sun, Zhuang Liu, Anna Bair, and J~Zico Kolter.
\newblock A simple and effective pruning approach for large language models.
\newblock \emph{arXiv preprint arXiv:2306.11695}, 2023.
\newblock URL \url{https://arxiv.org/pdf/2306.11695.pdf}.

\bibitem[Thakkar et~al.(2023)Thakkar, Ramani, Cecka, Shivam, Lu, Yan, Kosaian, Hoemmen, Wu, Kerr, Nicely, Merrill, Blasig, Qiao, Majcher, Springer, Hohnerbach, Wang, and Gupta]{Thakkar_CUTLASS_2023}
Vijay Thakkar, Pradeep Ramani, Cris Cecka, Aniket Shivam, Honghao Lu, Ethan Yan, Jack Kosaian, Mark Hoemmen, Haicheng Wu, Andrew Kerr, Matt Nicely, Duane Merrill, Dustyn Blasig, Fengqi Qiao, Piotr Majcher, Paul Springer, Markus Hohnerbach, Jin Wang, and Manish Gupta.
\newblock {CUTLASS}, Jan 2023.
\newblock URL \url{https://github.com/NVIDIA/cutlass}.

\bibitem[Wan et~al.(2023)Wan, Wang, Liu, Alam, Zheng, Liu, Qu, Yan, Zhu, Zhang, et~al.]{wan2023efficient}
Zhongwei Wan, Xin Wang, Che Liu, Samiul Alam, Yu~Zheng, Jiachen Liu, Zhongnan Qu, Shen Yan, Yi~Zhu, Quanlu Zhang, et~al.
\newblock Efficient large language models: A survey.
\newblock \emph{Transactions on Machine Learning Research}, 2023.
\newblock URL \url{https://openreview.net/pdf?id=bsCCJHbO8A}.

\bibitem[Wang et~al.(2024{\natexlab{a}})Wang, Gao, Zhao, Sun, and Dai]{wang2024auxiliary}
Lean Wang, Huazuo Gao, Chenggang Zhao, Xu~Sun, and Damai Dai.
\newblock Auxiliary-loss-free load balancing strategy for mixture-of-experts.
\newblock \emph{arXiv preprint arXiv:2408.15664}, 2024{\natexlab{a}}.
\newblock URL \url{https://arxiv.org/pdf/2408.15664}.

\bibitem[Wang et~al.(2024{\natexlab{b}})Wang, Chen, and Zhu]{wang2024remoe}
Ziteng Wang, Jianfei Chen, and Jun Zhu.
\newblock {ReMoE}: Fully differentiable mixture-of-experts with {ReLU} routing.
\newblock \emph{arXiv preprint arXiv:2412.14711}, 2024{\natexlab{b}}.
\newblock URL \url{https://arxiv.org/pdf/2412.14711}.

\bibitem[Wei et~al.(2024)Wei, Wang, Liu, Ding, and Zhang]{wei2024magicoder}
Yuxiang Wei, Zhe Wang, Jiawei Liu, Yifeng Ding, and Lingming Zhang.
\newblock Magicoder: Empowering code generation with {OSS-Instruct}.
\newblock In \emph{Forty-first International Conference on Machine Learning}, 2024.
\newblock URL \url{https://arxiv.org/pdf/2312.02120}.

\bibitem[Xia et~al.(2024)Xia, Yang, Dong, Wang, Li, Ge, Liu, Li, and Sui]{xia2024unlocking}
Heming Xia, Zhe Yang, Qingxiu Dong, Peiyi Wang, Yongqi Li, Tao Ge, Tianyu Liu, Wenjie Li, and Zhifang Sui.
\newblock Unlocking efficiency in large language model inference: A comprehensive survey of speculative decoding.
\newblock In \emph{Findings of the Association for Computational Linguistics ACL 2024}, pp.\  7655--7671, 2024.
\newblock URL \url{https://aclanthology.org/2024.findings-acl.456.pdf}.

\bibitem[Xia et~al.(2023)Xia, Gao, Zeng, and Chen]{xia2023sheared}
Mengzhou Xia, Tianyu Gao, Zhiyuan Zeng, and Danqi Chen.
\newblock Sheared {LLaMA}: Accelerating language model pre-training via structured pruning.
\newblock \emph{arXiv preprint arXiv:2310.06694}, 2023.
\newblock URL \url{https://arxiv.org/pdf/2310.06694.pdf}.

\bibitem[Xiao et~al.(2023)Xiao, Lin, Seznec, Wu, Demouth, and Han]{xiao2023smoothquant}
Guangxuan Xiao, Ji~Lin, Mickael Seznec, Hao Wu, Julien Demouth, and Song Han.
\newblock Smoothquant: Accurate and efficient post-training quantization for large language models.
\newblock In \emph{International Conference on Machine Learning}, pp.\  38087--38099. PMLR, 2023.
\newblock URL \url{https://proceedings.mlr.press/v202/xiao23c/xiao23c.pdf}.

\bibitem[Xu et~al.(2023)Xu, Sun, Zheng, Geng, Zhao, Feng, Tao, and Jiang]{xu2023wizardlm}
Can Xu, Qingfeng Sun, Kai Zheng, Xiubo Geng, Pu~Zhao, Jiazhan Feng, Chongyang Tao, and Daxin Jiang.
\newblock {WizardLM}: Empowering large language models to follow complex instructions.
\newblock \emph{arXiv preprint arXiv:2304.12244}, 2023.
\newblock URL \url{https://arxiv.org/pdf/2304.12244}.

\bibitem[Xue et~al.(2024)Xue, Song, Mi, Chen, Xia, and Chen]{xue2024powerinfer}
Zhenliang Xue, Yixin Song, Zeyu Mi, Le~Chen, Yubin Xia, and Haibo Chen.
\newblock {PowerInfer-2}: Fast large language model inference on a smartphone.
\newblock \emph{arXiv preprint arXiv:2406.06282}, 2024.
\newblock URL \url{https://arxiv.org/pdf/2406.06282}.

\bibitem[Yang et~al.(2022)Yang, Hu, Babuschkin, Sidor, Liu, Farhi, Ryder, Pachocki, Chen, and Gao]{yang2022tensor}
Greg Yang, Edward~J Hu, Igor Babuschkin, Szymon Sidor, Xiaodong Liu, David Farhi, Nick Ryder, Jakub Pachocki, Weizhu Chen, and Jianfeng Gao.
\newblock Tensor programs {V}: Tuning large neural networks via zero-shot hyperparameter transfer.
\newblock \emph{arXiv preprint arXiv:2203.03466}, 2022.
\newblock URL \url{https://arxiv.org/pdf/2203.03466}.

\bibitem[Yao et~al.(2023)Yao, Li, Wu, Youn, and He]{yao2023comprehensive}
Zhewei Yao, Cheng Li, Xiaoxia Wu, Stephen Youn, and Yuxiong He.
\newblock A comprehensive study on post-training quantization for large language models.
\newblock \emph{arXiv preprint arXiv:2303.08302}, 2023.
\newblock URL \url{https://arxiv.org/pdf/2303.08302.pdf}.

\bibitem[Zellers et~al.(2019)Zellers, Holtzman, Bisk, Farhadi, and Choi]{hellaswag}
Rowan Zellers, Ari Holtzman, Yonatan Bisk, Ali Farhadi, and Yejin Choi.
\newblock {HellaSwag}: Can a machine really finish your sentence?
\newblock In \emph{Proceedings of the 57th Annual Meeting of the Association for Computational Linguistics}, pp.\  4791--4800, 2019.
\newblock URL \url{https://aclanthology.org/P19-1472.pdf}.

\bibitem[Zhang \& Sennrich(2019)Zhang and Sennrich]{zhang2019root}
Biao Zhang and Rico Sennrich.
\newblock Root mean square layer normalization.
\newblock \emph{Advances in Neural Information Processing Systems}, 32, 2019.
\newblock URL \url{https://proceedings.neurips.cc/paper/2019/file/1e8a19426224ca89e83cef47f1e7f53b-Paper.pdf}.

\bibitem[Zhang et~al.(2024{\natexlab{a}})Zhang, Song, Yu, Han, Lin, Xiao, Song, Liu, Mi, and Sun]{zhang2024relu}
Zhengyan Zhang, Yixin Song, Guanghui Yu, Xu~Han, Yankai Lin, Chaojun Xiao, Chenyang Song, Zhiyuan Liu, Zeyu Mi, and Maosong Sun.
\newblock {ReLU}$^2$ wins: Discovering efficient activation functions for sparse {LLMs}.
\newblock \emph{arXiv preprint arXiv:2402.03804}, 2024{\natexlab{a}}.
\newblock URL \url{https://arxiv.org/pdf/2402.03804.pdf}.

\bibitem[Zhang et~al.(2024{\natexlab{b}})Zhang, Xiao, Qin, Lin, Zeng, Han, Liu, Xie, Sun, and Zhou]{zhang2024exploring}
Zhengyan Zhang, Chaojun Xiao, Qiujieli Qin, Yankai Lin, Zhiyuan Zeng, Xu~Han, Zhiyuan Liu, Ruobing Xie, Maosong Sun, and Jie Zhou.
\newblock Exploring the benefit of activation sparsity in pre-training.
\newblock In \emph{Forty-first International Conference on Machine Learning}, 2024{\natexlab{b}}.
\newblock URL \url{https://openreview.net/pdf?id=KfXXPCcobh}.

\bibitem[Zhao et~al.(2024)Zhao, Huang, Han, Xu, Xiao, Zhang, Fang, Zhang, Liu, and Sun]{zhao2024ouroboros}
Weilin Zhao, Yuxiang Huang, Xu~Han, Wang Xu, Chaojun Xiao, Xinrong Zhang, Yewei Fang, Kaihuo Zhang, Zhiyuan Liu, and Maosong Sun.
\newblock Ouroboros: Generating longer drafts phrase by phrase for faster speculative decoding.
\newblock In \emph{Proceedings of the 2024 Conference on Empirical Methods in Natural Language Processing}, pp.\  13378--13393, 2024.
\newblock URL \url{https://aclanthology.org/2024.emnlp-main.742.pdf}.

\bibitem[Zhao et~al.(2025)Zhao, Pan, Han, Zhang, Sun, Huang, Zhang, Zhao, Li, Wang, et~al.]{zhao2025fr}
Weilin Zhao, Tengyu Pan, Xu~Han, Yudi Zhang, Ao~Sun, Yuxiang Huang, Kaihuo Zhang, Weilun Zhao, Yuxuan Li, Jianyong Wang, et~al.
\newblock {FR-Spec}: Accelerating large-vocabulary language models via frequency-ranked speculative sampling.
\newblock \emph{arXiv preprint arXiv:2502.14856}, 2025.
\newblock URL \url{https://arxiv.org/pdf/2502.14856}.

\bibitem[Zhong et~al.(2024)Zhong, Xia, Chen, and Lewis]{zhong2024lory}
Zexuan Zhong, Mengzhou Xia, Danqi Chen, and Mike Lewis.
\newblock Lory: Fully differentiable mixture-of-experts for autoregressive language model pre-training.
\newblock \emph{arXiv preprint arXiv:2405.03133}, 2024.
\newblock URL \url{https://arxiv.org/pdf/2405.03133}.

\end{thebibliography}
\bibliographystyle{colm2025_conference}

\clearpage

\appendix
\section{Influence of Load Balancing} \label{sec:load-balancing}

As a common practice of training MoE models, load balancing is to make the activation frequency of each expert as balanced as possible so that the model can be more friendly for the distributed deployment with expert parallelism, where experts are separately deployed on different devices and work concurrently. However, most end-side devices (which this work mainly focuses on) do not contain so many computation devices or cores, and thus cannot well support expert parallelism. Instead, it is more important to reduce global computation costs and promote activation locality, which is critical for end-side deployment techniques such as offloading and speculative decoding.

\begin{table*}[ht]
    \footnotesize
    \centering
    \begin{tabular}{lccc}
    \toprule
    Setting & $TLS$ & $CLS_8\uparrow$ & PPL$\downarrow$ \\
    \midrule
    \textbf{AL+CS} & 80.54 & 71.38 & 14.88 \\
    \textbf{AL+CS+LB} & 78.06 & 69.68 & 15.26 \\
    \bottomrule
    \end{tabular}
    \caption{Load balancing is less important for end-side deployment and can cause potential performance degradation. ``LB'' indicates the load balancing with auxiliary loss.}
    \label{tab:loss-balancing}
\end{table*}

Therefore, in this work, we do not consider load balancing and only focus on sparsification and the activation locality issue. Besides, load balancing can potentially cause performance degradation~\citep{wang2024auxiliary}. Specifically, as shown in Table~\ref{tab:loss-balancing}, under similar TLS, the PPL suffers from a considerable increase after adding the load-balancing auxiliary loss.

\section{Adaptive Factor Scheduler for Chunk Sparsification Loss} \label{sec:adaptive-scheduler}

With the language modeling loss $\mathcal{L}_{lm}$, the training objective is:
\begin{small}
\begin{equation}
    \begin{aligned}
    \label{eq:overall-objective}
    \mathcal{L}_{total}=\mathcal{L}_{lm}+\lambda_{al}\mathcal{L}_{al}+\lambda_{cs}\mathcal{L}_{cs},
    \end{aligned}
\end{equation}
\end{small}
\hspace{-\fontdimen2\font}where $\lambda_{al}$ and $\lambda_{cs}$ are corresponding factors.

Considering the difficulty of tuning hyper-parameters, we introduce an adaptive factor scheduler for $\lambda_{cs}$, which controls the overall sparsity level. Concretely, this scheduler keeps $\lambda_{cs}$ constant as the initial value $\lambda^0_{cs}$ for the first $N_{st}$ steps. Next, for every $N_{adj}$ steps, the scheduler adjusts $\lambda_{cs}$ according to the change of $\mathcal{L}_{cs}$, increasing the factor when $\mathcal{L}_{cs}$ increases and vice versa. Formally, the behavior of this scheduler at step $m=(i+1) N_{adj}$ is:
\begin{small}
\begin{equation}
\begin{aligned}
    \label{eq:adaptive-scheduler}
    \lambda^{i+1}_{cs}&=
    \begin{cases}
    \lambda^0_{cs} & \mathrm{if}\ m\leq N_{st}\\
    \gamma_{cs}\cdot \lambda^{i}_{cs} & \mathrm{else\ if}\ \gamma_{cs}\leq1\\
    \max(\gamma_{min},\gamma_{cs})\cdot \lambda^{i}_{cs} & \mathrm{otherwise}
    \end{cases}\quad
    &\gamma_{cs}=\frac{\mathrm{Avg}[\mathcal{L}^t_{cs}]_{t=i\cdot N_{adj}}^{(i+1) N_{adj}}}{\mathrm{Avg}[\mathcal{L}^t_{cs}]_{t=(i-1)N_{adj}}^{i\cdot N_{adj}}},\\
\end{aligned}
\end{equation}
\end{small}
\hspace{-\fontdimen2\font}where $\mathcal{L}^t_{cs}$ denotes the loss value at step $t$, and $\gamma_{min}$ is the minimum magnification ratio.

\section{Experimental Settings} \label{sec:experiment-setting}

First, we give a detailed introduction to baseline MoE architectures used in our experiment:

(1) \textbf{Vanilla TopK MoE} is currently the most common MoE implementation, adopted by works such as Switch Transformer~\citep{fedus2022switch} and Mixtral~\citep{jiang2024mixtral}. Their routers are composed of the softmax and TopK functions.

(2) \textbf{DeepSeekMoE (DSMoE)}~\citep{dai2024deepseekmoe} adopts a similar TopK MoE architecture but introduces shared experts for improvement, which are consistently activated by each token.

(3) \textbf{GRIN}~\citep{liu2024grin}, also using the TopK activation, adopts an innovative routing strategy called SparseMixer-v2. This alleviates the non-differentiable issue through an approximation of the missing gradients.

(4) \textbf{ReMoE}~\citep{wang2024remoe} makes the router differentiable by introducing a ReLU-based router module. Though similar to our design, ReMoE does not apply RMSNorm after ReLU, and more importantly, its L1 regularization directly imposed on activation values can cause activation magnitude disturbance (Section~\ref{sec:rmsnorm-disturbance}) and harm performance.

\begin{table*}[t]
    \footnotesize
    \centering
    \begin{tabular}{lccccccccc}
    \toprule
    Scale & $d_h$ & $d_e$ & $N_e$ & $N_{layer}$ & $N_{tot}$ & $\lambda_{al}$ & $\lambda_{cs}^0$ & $batch\ size$ & $n_{pre}$\\
    \midrule
    0.1B (Small) & 768 & 64 & 48 & 16 & $1.19\times10^8$ & $2e-3$ & $5e-2$ & $1.57\times10^6$ & 10000 \\
    0.5B (Medium) & 1280 & 128 & 40 & 27 & $4.95\times10^8$ & $2e-3$ & $5e-2$ & $3.15\times10^6$ & 20000 \\
    0.8B (Large) & 1536 & 128 & 48 & 32 & $8.17\times10^8$ & $1e-3$ & $5e-2$ & $2.36\times10^6$ & 30000 \\
    1.2B (XLarge) & 1792 & 128 & 56 & 35 & $1.19\times10^9$ & $2e-3$ & $5e-2$ & $3.15\times10^6$ & 40000 \\
    2.8B & 2048 & 128 & 128 & 36 & $2.80\times10^9$ & $2e-3$ & $1e-1$ & $3.15\times10^6$ & 25000 \\
    \bottomrule
    \end{tabular}
    \caption{The major structural settings and hyper-parameters of our experimental models. $N_{layer}$, $N_{tot}$, and $n_{pre}$ denote the number of layers, the number of non-embedding parameters, and the pre-training steps, respectively.}
    \label{tab:model-setting}
    \vspace{-1em}
\end{table*}

Next, we list the hyper-parameters in Table~\ref{tab:model-setting}. We adopt $\mu P$ parametrization~\citep{yang2022tensor} to promote training stability and reduce the influence of hyper-parameters. Therefore, we can adopt the same setting for the following parameters: peak learning rate $lr=0.01$, $\beta_1=0.9$, $\beta_2=0.95$, $weight\ decay = 0.1$. We use the WSD scheduler to adjust the learning rates in the training process~\citep{hu2024minicpm,dubey2024llama}. As for the adaptive factor scheduler, under all BlockFFN settings, we adjust the factor every $N_{adj}=100$ steps, with $N_{st}=1000$ and $\gamma_{min}=1.025$.

\section{Model Settings} \label{sec:model-setting}

For all settings, which include BlockFFN, the upper bound ``Dense'', and the other MoE baselines, we maintain the close number of total parameters, activated parameters (i.e., TLS), and training tokens. Moreover, the number and the intermediate dimension of experts are also exactly the same, following the fine-grained expert segmentation of DeepSeekMoE~\citep{liu2024deepseek}. As for the attention layer, we apply the multi-latent attention (MLA)~\citep{liu2024deepseekv2} for models from 0.1B to 1.2B, while adopting group query attention (GQA) for BlockFFN-2.8B to make the acceleration implementation easier. Therefore, we ensure that the differences between different settings only lie in the routing strategy and training objectives, which are the key improved points of our work. The detailed structural settings of our models are listed in Table~\ref{tab:model-setting}.

\section{Datasets and Benchmarks} \label{sec:data-benchmark}

\paragraph{Training data} The pre-training data of BlockFFN is a comprehensive mixture of multiple corpora across various categories. This includes C4~\citep{raffel2020exploring}, Pile~\citep{gao2020pile}, Dolma~\citep{soldaini2024dolma}, CommonCrawl, StarCoder~\citep{li2023starcoder}, and other collected raw corpus. Besides, to obtain reasonable evaluation results, we perform a decay stage before evaluating models on benchmarks~\citep{hu2024minicpm,dubey2024llama}. For this stage, instruction-tuning data are added, including EvolInstruct~\citep{xu2023wizardlm}, UltraChat~\citep{ding2023enhancing}, OssInstruct~\citep{wei2024magicoder}, and other collected SFT datasets.

\paragraph{Validation data} The validation data has the same distribution as the pre-training data. Deduplication is conducted to alleviate the intersections between pre-training and validation data, so that the validation data cannot be easily over-fitted.

\paragraph{Evaluation benchmarks} The task-specific benchmarks used in our experiments can be divided into two groups: commonsense reasoning (C.R.) and reading comprehension (R.C.). The former group includes PIQA~\citep{piqa}, SIQA~\citep{siqa}, and HellaSwag~\citep{hellaswag}. The latter group includes LAMBADA~\citep{lambada}, TyDi QA~\citep{tydiqa}, and BoolQ~\citep{boolq}. For both groups, the evaluation metric is 0-shot accuracy.

\paragraph{Performance on independent benchmarks} In Table~\ref{tab:moe-eval}, we only list the average evaluation scores of two benchmark groups. In this section, we provide the evaluation results on independent benchmarks, as shown in Table~\ref{tab:eval-cr} and~\ref{tab:eval-rc}.

\begin{table*}[t]
    \footnotesize
    \centering
    \setlength{\tabcolsep}{0.3em}
    \begin{tabular}{lccc|ccc|ccc|ccc}
    \toprule
    \multirow{2}{*}{Setting} & \multicolumn{3}{c|}{Small} & \multicolumn{3}{c|}{Medium} & \multicolumn{3}{c|}{Large} & \multicolumn{3}{c}{XLarge} \\
    \cmidrule{2-13}
    & PIQA & SIQA & Hella. & PIQA & SIQA & Hella. & PIQA & SIQA & Hella. & PIQA & SIQA & Hella. \\
    \midrule
    Dense & 65.51 & 38.23 & 31.71 & 69.80 & 40.53 & 47.71 & 70.62 & 42.27 & 51.77 & 71.98 & 42.32 & 57.61 \\
    \midrule

    TopK & 63.98 & 36.28 & 30.93 & 68.44 & 40.43 & 43.82 & 69.53 & 41.45 & 47.55 & 71.60 & 41.86 & 51.49 \\
    DSMoE & 65.13 & 36.95 & 30.96 & 68.28 & 40.02 & 43.09 & 69.31 & 40.99 & 47.33 & 70.84 & 41.61 & 53.40 \\
    GRIN & 64.20 & 36.08 & 30.78 & 68.28 & 40.02 & 44.59 & 69.31 & 41.35 & 48.13 & 69.86 & 41.76 & 52.34 \\
    ReMoE & 64.91 & 38.18 & 32.56 & 69.31 & 40.38 & 44.57 & 71.38 & 40.38 & 49.55 & 71.06 & 42.22 & 53.40 \\

    \midrule

    \textbf{BlockFFN} & 64.96 & 37.62 & 31.82 & 69.80 & 39.56 & 45.89 & 71.49 & 41.40 & 50.43 & 71.16 & 42.84 & 55.26 \\
    \bottomrule
    \end{tabular}
    \caption{The evaluation scores on the three benchmarks of commonsense reasoning (C.R.).}
    \label{tab:eval-cr}
    \vspace{-1em}
\end{table*}

\begin{table*}[t]
    \footnotesize
    \centering
    \setlength{\tabcolsep}{0.3em}
    \begin{tabular}{lccc|ccc|ccc|ccc}
    \toprule
    \multirow{2}{*}{Setting} & \multicolumn{3}{c|}{Small} & \multicolumn{3}{c|}{Medium} & \multicolumn{3}{c|}{Large} & \multicolumn{3}{c}{XLarge} \\
    \cmidrule{2-13}
    & LAM. & TyDi. & BoolQ & LAM. & TyDi. & BoolQ & LAM. & TyDi. & BoolQ & LAM. & TyDi. & BoolQ \\
    \midrule

    Dense & 30.41 & 13.41 & 54.34 & 43.80 & 36.59 & 54.16 & 49.04 & 39.77 & 57.80 & 52.22 & 40.23 & 56.15 \\
    \midrule

    TopK & 29.03 & 5.91 & 52.02 & 42.40 & 34.55 & 50.46 & 44.85 & 41.59 & 42.72 & 49.16 & 40.23 & 58.47 \\
    DSMoE & 28.37 & 17.05 & 50.28 & 43.74 & 33.41 & 57.52 & 48.52 & 44.77 & 46.06 & 50.30 & 47.95 & 53.46 \\
    GRIN & 28.41 & 12.27 & 54.16 & 42.77 & 35.00 & 49.91 & 44.87 & 38.18 & 52.08 & 50.44 & 44.55 & 54.04 \\
    ReMoE & 29.44 & 15.68 & 53.21 & 42.58 & 32.05 & 54.53 & 45.90 & 39.55 & 56.54 & 51.35 & 30.23 & 59.79 \\

    \midrule

    \textbf{BlockFFN} & 30.53 & 21.14 & 50.12 & 42.75 & 37.50 & 50.98 & 45.97 & 46.59 & 62.23 & 50.30 & 43.41 & 58.47 \\
    \bottomrule
    \end{tabular}
    \caption{The evaluation scores on the three benchmarks of reading comprehension (R.C.).}
    \label{tab:eval-rc}
    \vspace{-1em}
\end{table*}

\section{Effect of Expert Granularity} \label{sec:expert-granularity}

\begin{figure}[ht]
\begin{minipage}{0.48\textwidth}
    \centering
    \includegraphics[width=\linewidth]{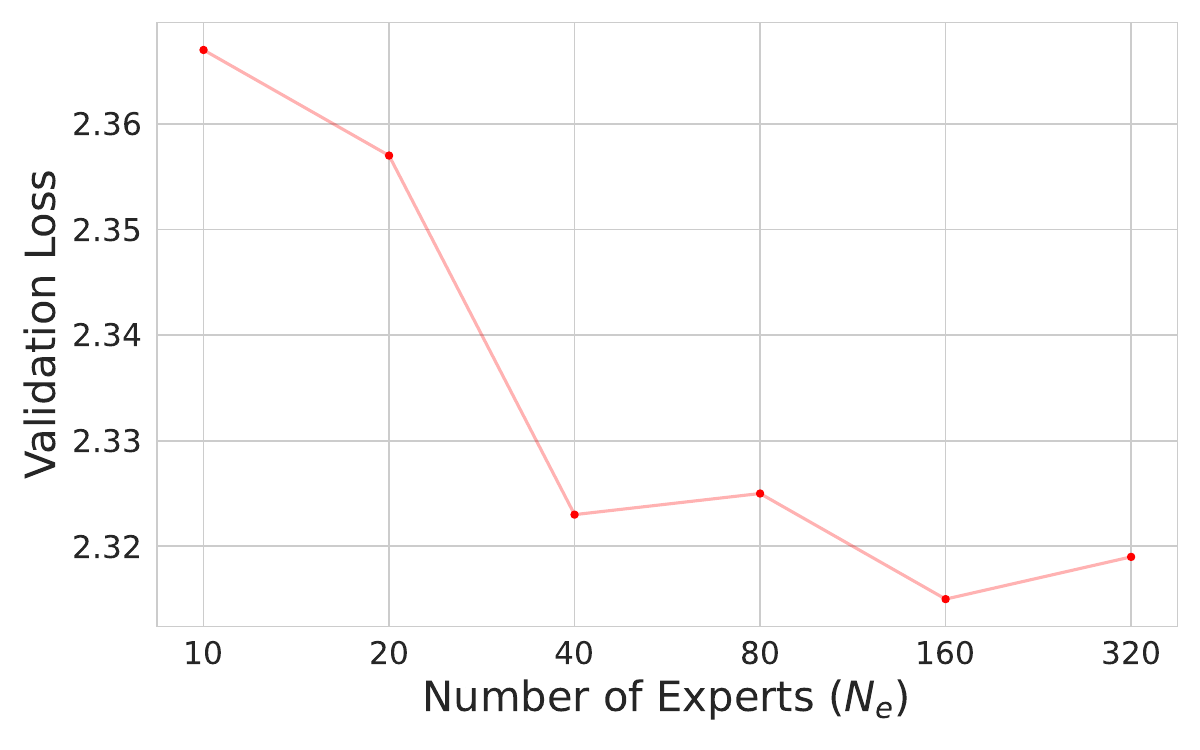}
    \caption{The validation loss of BlockFFN Medium with different expert granularities.}
    \label{fig:granularity-loss}
\end{minipage}
\hfill
\begin{minipage}{0.48\textwidth}
    \centering
    \includegraphics[width=\linewidth]{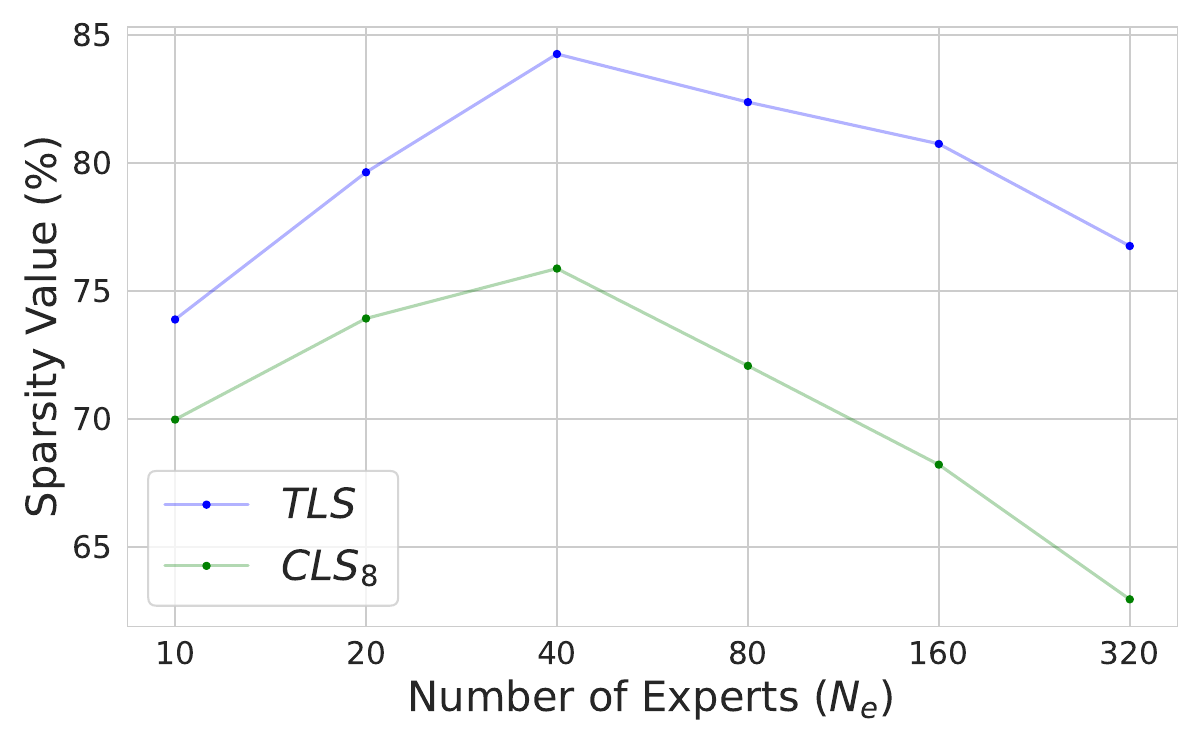}
    \caption{The TLS and CLS of BlockFFN Medium with different expert granularities.}
    \label{fig:granularity-sparsity}
\end{minipage}
\end{figure}

Expert granularity has long been demonstrated to influence the performance of MoE models~\citep{krajewski2024scaling}. Specifically, \textbf{given a fixed computation budget (assumed proportional to the parameter scale), what is the best trade-off between the expert number $N_e$ and expert dimension $d_e$}? To solve this problem, we conduct experiments on BlockFFN Medium with different expert granularities. These models are evaluated from four aspects: the validation loss, the token-level sparsity, the chunk-level sparsity, and memory locality.

First, as shown in Figure~\ref{fig:granularity-loss}, while the loss drops considerably with coarse granularities (i.e., small $N_e$), the marginal benefits of granularity increase gradually diminishes with $>40$ experts. Moreover, Figure~\ref{fig:granularity-sparsity} further displays a nonmonotonic relationship between sparsity and expert granularities. The setting with 40 experts, which we adopt in our main experiments (Section~\ref{sec:overall-result}), achieves both the highest $TLS$ and $CLS_8$. Finally, as larger memory access units generally have better memory locality and hardware-friendliness, we do not expect an extremely fine granularity. To sum up, $40$ experts is the best setting for BlockFFN Medium. We leave more quantitative analyses for future studies.

\section{Upper Bound Analysis of Acceleration Kernels} \label{sec:upper-bound}

\begin{table*}[ht]
    \footnotesize
    \centering
    \setlength{\tabcolsep}{0.4em}
    \begin{tabular}{lcc|lcc}
    \toprule
    Setting & FFN Time (ms) & $TLS$ & Setting & FFN Time (ms) & $CLS_{spec}$ \\
    \midrule
    Baseline AR & 61 & \multirow{2}{*}{$1-\mathbf{12.45\%}$} & EAGLE-2 & 88 & \multirow{2}{*}{$1-\mathbf{30.86\%}$}\\
    \textbf{Ours (1-Tok)} & 7.8 ($\mathbf{12.8\%}\times$) & & \textbf{Ours (32-Tok)} & 27 ($\mathbf{30.7\%}\times$) & \\
    \bottomrule
    \end{tabular}
    \caption{The upper bound analysis of our kernels. $TLS$ and $CLS_{spec}$ values are evaluated on Spec-Bench decoding tokens, which are close to the FFN time consumption ratios of ``Ours (1-Tok) / Baseline AR'' and ``Ours (32-Tok) / EAGLE-2'', respectively.}
    \label{tab:upper-bound}
\end{table*}

To delve deep into the ability of our acceleration kernels, we conduct an upper bound analysis by inspecting the time consumption of FFNs separately. As shown in Table~\ref{tab:upper-bound}, ``Baseline AR'' and ``EAGLE-2'' can be viewed as the sparsity-ablation setting of ``Ours (1-Tok)'' and ``Ours (32-Tok)'', respectively, with only about 12.8\% and 30.7\% FFN time consumption. Surprisingly, we find that these two time consumption ratios are quite approximate to the $TLS$ and $CLS_{spec}$, respectively. Note that ``Ours (32-Tok)'' adopts a draft tree size of 32, and $CLS_{spec}$ is calculated by the average ratio of experts activated by the union of all the 32 tokens contained in the EAGLE-2 draft tree.
This phenomenon indicates that both kernels can reach the theoretical speedup upper bound in FFN acceleration induced by the corresponding token-level sparsity and the union sparsity of tokens in a draft tree.

Notably, although our CLS-aware training objectives do not directly optimize the tree-level union sparsity, these objectives tailored for consecutive chunks are effective for tree patterns, since each path from the root node to the leaf node is still composed of consecutive tokens.

\section{Inference Acceleration on Independent Datasets} \label{sec:speedup-independent}

\begin{table*}[ht]
    \footnotesize
    \centering
    \setlength{\tabcolsep}{0.3em}
    \begin{tabular}{lcc|cc|cc|cc}
    \toprule
    \multirow{2}{*}{Setting} & \multicolumn{2}{c|}{\textbf{MT.}} & \multicolumn{2}{c|}{\textbf{Trans.}} & \multicolumn{2}{c|}{\textbf{Summ.}} & \multicolumn{2}{c}{\textbf{QA}} \\
    & Tokens/s & Speedup & Tokens/s & Speedup & Tokens/s & Speedup & Tokens/s & Speedup \\
    \midrule
    Huggingface & 7.61 & 0.60 & 7.32 & 0.55 & 7.54 & 0.61 & 7.81 & 0.60 \\
    \midrule
    Baseline AR & 12.61 & 1.00 & 13.32 & 1.00 & 12.40 & 1.00 & 13.04 & 1.00 \\
    EAGLE-2 & 23.70 & 1.88 & 20.26 & 1.52 & 20.74 & 1.67 & 21.99 & 1.69 \\
    \midrule
    \textbf{Ours (1-Tok)} & 38.41 & 3.05 & \textbf{40.39} & \textbf{3.03} & 37.37 & 3.01 & 42.05 & 3.22 \\
    \textbf{Ours (32-Tok)} & \textbf{49.43} & \textbf{3.92} & 39.18 & 2.94 & \textbf{42.70} & \textbf{3.44} & \textbf{46.09} & \textbf{3.53} \\
    \bottomrule
    \end{tabular}
    \caption{Detailed speedup results on NVIDIA Jetson Orin NX (1st part).}
    \label{tab:speedup-1}
\end{table*}

\begin{table*}[ht]
    \footnotesize
    \centering
    \setlength{\tabcolsep}{0.5em}
    \begin{tabular}{lcc|cc|cc}
    \toprule
    \multirow{2}{*}{Setting} & \multicolumn{2}{c|}{\textbf{Math}} & \multicolumn{2}{c|}{\textbf{RAG}} & \multicolumn{2}{c}{\textbf{Average}} \\
    & Tokens/s & Speedup & Tokens/s & Speedup & Tokens/s & Speedup \\
    \midrule
    Huggingface & 7.76 & 0.61 & 5.96 & 0.46 & 7.33 & 0.57 \\
    \midrule
    Baseline AR & 12.80 & 1.00 & 12.86 & 1.00 & 12.84 & 1.00 \\
    EAGLE-2 & 25.01 & 1.95 & 22.47 & 1.75 & 22.36 & 1.74 \\
    \midrule
    \textbf{Ours (1-Tok)} & 44.54 & 3.48 & 39.38 & 3.06 & 40.36 & 3.14 \\
    \textbf{Ours (32-Tok)} & \textbf{59.85} & \textbf{4.68} & \textbf{45.76} & \textbf{3.56} & \textbf{47.17} & \textbf{3.67} \\
    \bottomrule
    \end{tabular}
    \caption{Detailed speedup results on NVIDIA Jetson Orin NX (2nd part).}
    \label{tab:speedup-2}
\end{table*}

\begin{table*}[ht]
    \footnotesize
    \centering
    \begin{tabular}{cccccc|c}
    \toprule
    \textbf{MT.} & \textbf{Trans.} & \textbf{Summ.} & \textbf{QA} & \textbf{Math} & \textbf{RAG} & \textbf{Average} \\
    \midrule
    2.73 & \textbf{2.17} & 2.38 & 2.67 & 2.83 & 2.57 & 2.66 \\
    \bottomrule
    \end{tabular}
    \caption{The acceptance lengths on each independent dataset of Spec-Bench.}
    \label{tab:accept-length}
\end{table*}

In Table~\ref{tab:speedup-overall}, we provide the decoding speeds on each dataset contained in Spec-Bench and the average speedup ratio. In this section, we list the speedup ratios on each independent dataset, as shown in Table~\ref{tab:speedup-1} and~\ref{tab:speedup-2}.

On most datasets, ``Ours (32-tok)'' achieves the best inference efficiency. However, there exists an exception, ``Translation'' (\textbf{Trans.}), where ``Ours (32-tok)'' underperforms ``Ours (1-tok)''. This indicates that the combination of chunk-level activation sparsity and speculative decoding has worse performance than utilizing token-level activation sparsity alone. After careful examination, we find this is attributed to the shortest EAGLE-2 acceptance length on this dataset (Table~\ref{tab:accept-length}), which hurts the efficiency of speculative decoding. Therefore, the sparsity-involved speculative decoding is more reasonable when speculative decoding works efficiently, generally with longer acceptance lengths and larger models.

\section{Ablation Studies on the Gated Expert Variant} \label{sec:gated-expert}

For the design of BlockFFN expert modules, we choose the non-gated MLP instead of the more widely adopted gated variant. To support this choice, we conduct an ablation study on BlockFFN (Small). As shown in Table~\ref{tab:gated-expert}, using a gated MLP for expert modules can cause extremely low sparsity, which is quite a surprising result worth further study. A possible explanation may lie in the ``competition'' of sparsity between the router module and the expert module, indicating that the router sparsity and the expert sparsity vary in the opposite direction. Therefore, the higher sparsity of gated MLPs can significantly weaken the sparsity of the router module.

\begin{table*}[ht]
    \footnotesize
    \centering
    \begin{tabular}{l|cccc}
    \toprule
    Expert Design & $TLS\uparrow$ & $CLS_8\uparrow$ & PPL$\downarrow$ \\
    \midrule
    Non-gated & 80.54 & 71.38 & 14.88 \\
    Gated & 28.39 & 25.44 & 15.17 \\
    \bottomrule
    \end{tabular}
    \caption{The ablation results of the gated variant for BlockFFN expert modules.}
    \label{tab:gated-expert}
    \vspace{-1.5em}
\end{table*}

\section{More Details about Acceleration Kernels} \label{sec:kernel-detail}

\begin{figure}[ht]
    \centering
    \includegraphics[width=0.9\linewidth]{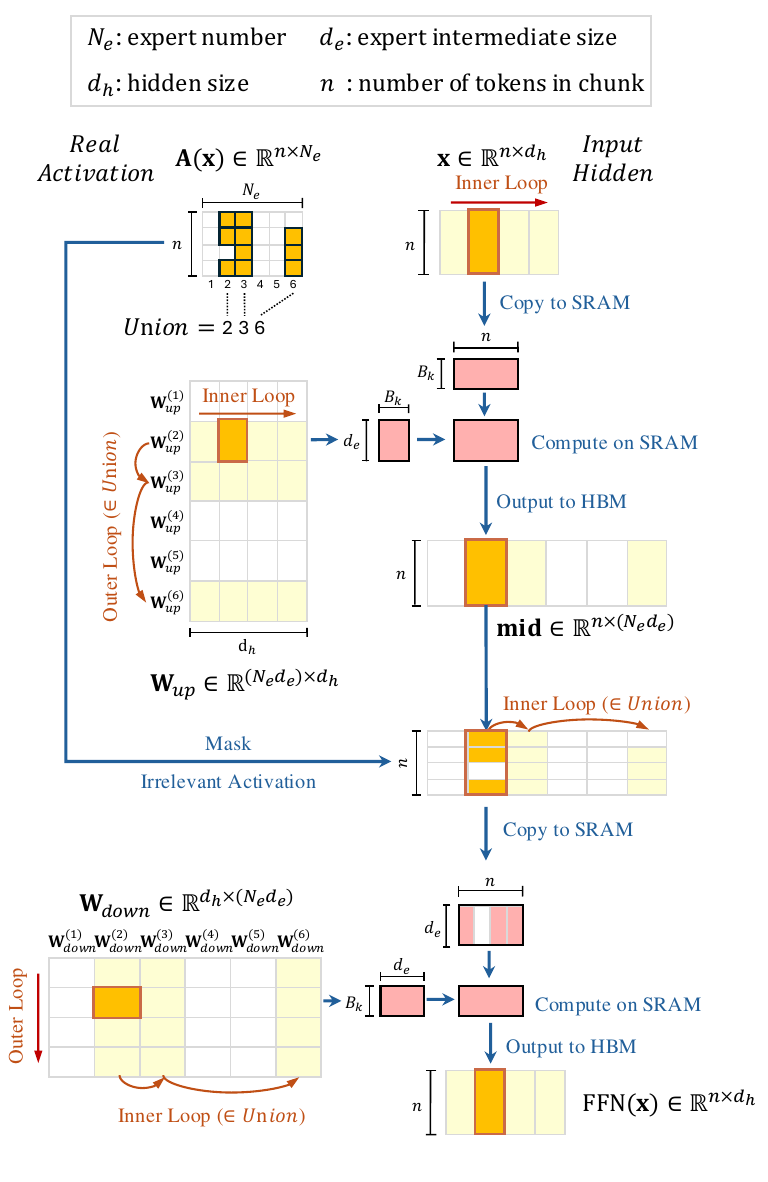}
    \caption{The detailed framework of our efficient acceleration kernels.}
    \label{fig:blockffn-kernel-detail}
\end{figure}

Figure~\ref{fig:blockffn-kernel-detail} shows more details about our efficient acceleration kernels. These include the execution details of the two-loop structure of the kernels, and how the weights of activated experts are transferred between different memory hardware (e.g., SRAM and HBM).

\end{document}